\documentclass[letterpaper]{article}
\usepackage[draft]{aaai2026}
\usepackage{times}
\usepackage{helvet}
\usepackage{courier}
\usepackage[hyphens]{url}
\usepackage{graphicx}
\usepackage{natbib}
\usepackage{caption}
\usepackage{booktabs}
\usepackage{amsmath}
\usepackage{amssymb}
\usepackage{algorithm}
\usepackage{algpseudocode}
\usepackage{multirow}
\usepackage{colortbl}
\usepackage{enumitem}
\usepackage{subcaption}

\ifdefined
\fi

\newcommand{\eg}{e.g.}

\newcommand{\mypara}[1]{\smallskip\noindent\textbf{#1}\enspace}
\newcommand{\keywords}[1]{\par\noindent\textbf{Keywords: }{\def\and{; }#1}}

\title{PrecogUI: Proactive GUI Agents via\\Pre-cognitive Simulation and Experience Retrieval}
\author{
Bin Kang\textsuperscript{\rm 1,\rm 3,\rm 5},
Jiarui Ouyang\textsuperscript{\rm 2},
Li Jiang\textsuperscript{\rm 4},\\
Bin Chen\textsuperscript{\rm 3},
Zhuotao Tian\thanks{Corresponding author.}\textsuperscript{\rm 3,\rm 5}
}
\affiliations{
\textsuperscript{\rm 1}University of Chinese Academy of Sciences\\
\textsuperscript{\rm 2}The Hong Kong University of Science and Technology\\
\textsuperscript{\rm 3}Harbin Institute of Technology\\
\textsuperscript{\rm 4}The Chinese University of Hong Kong\\
\textsuperscript{\rm 5}Shenzhen Loop Area Institute
}

\begin{document}

\maketitle

\begin{abstract}
Existing reactive Graphical User Interface (GUI) agents often fail in long-horizon, dynamic scenarios, where unexpected disturbances trigger attention-diverting and cascading failures. To address this, we propose \textbf{PrecogUI}, a pre-cognitive architecture that shifts the paradigm from reactive execution to proactive decision-making. Specifically, we design a Proactive Experience Pool (PEP), which caches recurring anomaly and success patterns as "state-action-result" tuples in a dual-memory repository. Furthermore, we introduce a Proactive Simulation Executor (PSE) that learns to forecast the next symbolic UI layout given a candidate action, enabling early anomaly avoidance and ranking candidate actions by predicted reliability. Finally, a Pre-cognitive Execution Controller (PEC) fuses these priors and predictions, prioritizes handling of foreseen anomalies, and ensures execution robustness through a closed-loop error correction mechanism. For robust evaluation, we develop AutoTraj, an automatic data-generation engine, to construct InterfereBench, a benchmark for long-horizon tasks with strong disturbances. Experiments demonstrate that PrecogUI surpasses state-of-the-art methods on InterfereBench while maintaining competitive performance on public benchmarks. The code will be publicly available.
\end{abstract}

\section{Introduction}
\label{sec:intro}

Graphical User Interface (GUI) agents \cite{Cheng2024SeeClick, Lin_2025_CVPR, gou2025navigating, Hong_2024_CVPR} are built on Multimodal Large Language Models (MLLMs) to comprehend user queries, interpret context, and perform actions like clicks and swipes for accomplishing GUI tasks.
The advancement of MLLMs \cite{Li2023BLIP2, Alayrac2022Flaming, Dai2023InstructBLIP} has notably enhanced agents' interface perception and decision-making precision. Nevertheless, anomalies such as pop-ups and black screens in dynamic settings remain a significant hurdle, diverting attention and causing persistent cascading failures.

Prior research~\cite{Hong_2024_CVPR, Huang_2025_CVPR, Chen_2025_ACL} has significantly advanced the perception-action loop. However, the prevailing approach remains reactive, relying on current observations for decision-making. While effective in short-horizon, disturbance-free settings~\cite{rawles2025androidworld, NEURIPS2023_5950bf29}, these reactive methods may struggle in long-horizon tasks and dynamic environments. Recent efforts have attempted to address this challenge through online exploration~\cite{Sun_2025_CVPR, fan2025guibeealignguiaction} and app-specific memory or layout-aware retrieval~\cite{AutoDroid_AICMCN_2025, MapAgent_kong_2025}. Nevertheless, the reactive nature still leaves agents vulnerable to distractions from non-goal cues such as pop-ups and delays.

\begin{figure*}[t]
\centering
\includegraphics[width=.95\textwidth]{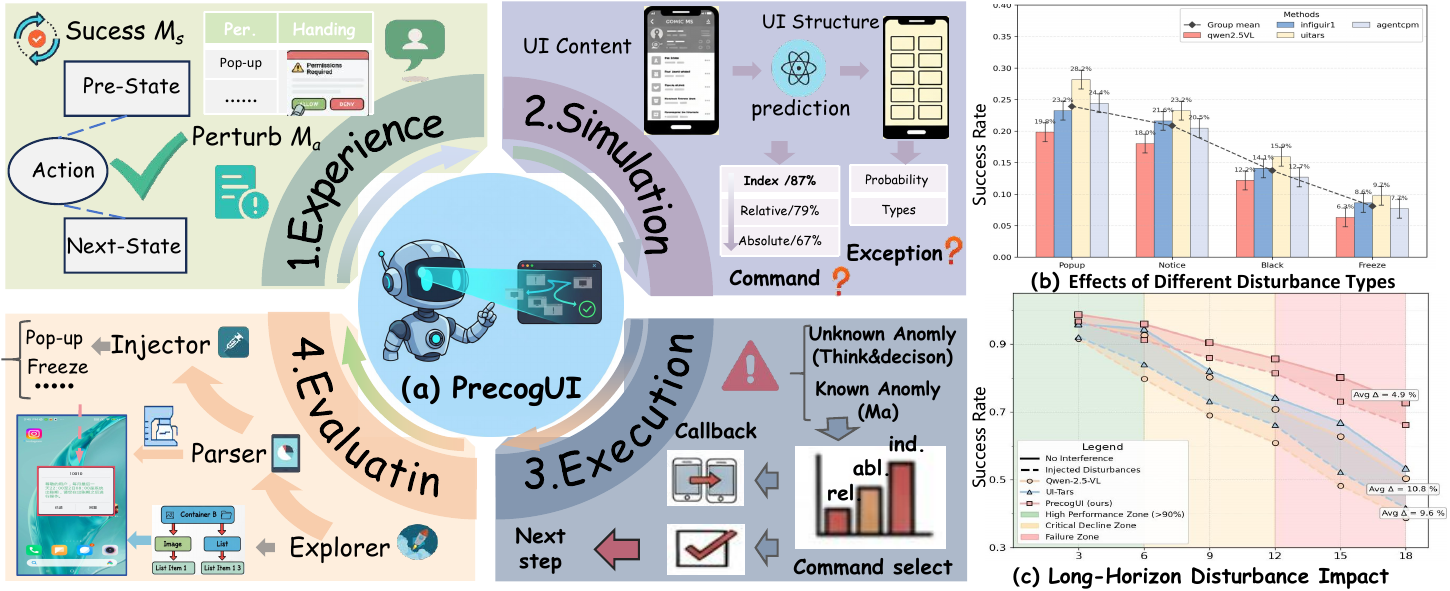} 
\caption{(a) PrecogUI combines experience retrieval with look-ahead simulation; AutoTraj generates perturbed trajectories. (b) Absolute SR drop by disturbance type on AndroidControl. (c) Disturbance impact grows with task horizon.}
\label{fig:intro}
\end{figure*}

\mypara{Key Observations.}
To investigate robustness, we evaluate representative reactive agents~\cite{liu2025_infiguir1, qin2025_uitars, zhang2025_agentcpm} on AndroidControl~\cite{NEURIPS2024_androidcontrol} under injected disturbances at both the overlay level (\eg, pop-ups, notifications) and environment level (\eg, black screens, freezing). Performance is assessed by success rate (SR), stratified by disturbance type and task horizon.
Specifically, the bars in Figure~\ref{fig:intro}(b) report absolute SR reductions by disturbance type. Overlay-level disturbances induce the most significant degradation, reducing SR by more than 20 percentage points on average, compared with an approximately 10-point drop under environment-level perturbations.
Besides, the performance degradation scales monotonically with horizon length, as shown in Figure~\ref{fig:intro}(c).
On short-horizon tasks ($<5$ steps), all models maintain high robustness (SR $\geq 91\%$). However, for medium-length tasks (6--15 steps), reactive agents exhibit increasingly pronounced SR degradation.
In long-horizon tasks ($>$15 steps), reactive agents' SR declines to approximately $50\%$.
See Appendix~\ref{sec:motivation} for further analysis.

These results show that reactive agents are easily distracted by non-goal stimuli, allowing errors to accumulate into cascading failures. This observation prompts a crucial question: \textit{how can we empower agents with pre-cognitive planning and explicit exception handling to ensure robustness in long-horizon, dynamic environments?}

\mypara{Our Solution.}
In this study, we propose PrecogUI, a framework that integrates experience retrieval with online look-ahead simulation to improve the robustness of GUI agents in long-horizon and disturbance-prone settings. The conceptual architecture is illustrated in Figure \ref{fig:intro}(a).

Specifically, PrecogUI introduces the Proactive Experience Pool (PEP), a dual-memory repository that stores and retrieves recurring interaction patterns from both successful and anomalous executions, enabling knowledge reuse via pattern matching. Then, the Proactive Simulation Executor (PSE) employs a conditional diffusion model \cite{Rombach_2022_CVPR} to simulate the symbolic UI layout resulting from candidate actions, providing look-ahead forecasts for early anomaly detection and candidate-action reliability ranking. Finally, these are integrated by the Pre-cognitive Execution Controller (PEC), which prioritizes anomaly handling, selects high-utility actions, and ensures robustness through state monitoring and hierarchical rollback/retry.

To the best of our knowledge, no existing benchmark systematically evaluates long-horizon robustness under diverse, sustained perturbations. We thus introduce InterfereBench, a new benchmark consisting of 1,160 task-level trajectory groups (${\sim}$27k annotated source screenshots) across 34 diverse applications. Each group provides a clean execution and two controlled perturbation replays, enabling paired evaluation under prolonged task horizons and dynamic interference. AutoTraj is the automated engine used to generate these perturbation-rich interaction trajectories at scale.
Experiments on InterfereBench and public benchmarks such as AndroidControl~\cite{NEURIPS2024_androidcontrol} and GUI-Odyssey~\cite{2024_gui_odyssey} show that PrecogUI outperforms the strongest baseline by 22.4 percentage points in success rate under strong perturbations, improving robustness without sacrificing overall performance.
To summarize, our contributions are as follows:
\begin{itemize}[leftmargin=*,topsep=0pt]
    \item We propose PrecogUI, a unified framework that combines offline experience reuse, proactive layout prediction, and exception-aware execution recovery to enhance robustness in long-horizon GUI interactions.
    \item We present InterfereBench, a new benchmark designed to evaluate robustness under strong, sustained perturbations in long-horizon tasks, along with AutoTraj, an automated pipeline for scalable, realistic trajectory generation.
    \item Extensive experiments on InterfereBench and the public benchmarks demonstrate that PrecogUI effectively improves long-horizon reliability and anomaly resilience while maintaining general GUI capabilities.
\end{itemize}


\section{Method}

\subsection{Overview}
\label{sec:overview}
Toward robust long-horizon execution under perturbations, we propose PrecogUI, which closes the loop between experience, foresight, and feedback via four modules:  (i) AutoTraj builds InterfereBench, a long-horizon benchmark with controlled perturbations; (ii) PEP forms a memory of anomaly/success patterns by indexing experiences based on their UI layout structure, enabling efficient retrieval of similar past cases; (iii) PSE predicts the next symbolic UI layout, estimates anomaly risk, and ranks candidate actions across index, relative, and absolute-level variants; (iv) PEC fuses PEP and PSE with online monitoring and rollback/retry to deliver robust, closed-loop control. We discuss related work in Sec.~\ref{sec:related_work_main}, with an extended review in Appendix~\ref{sec:Related_work}.

\subsection{Data Construction}\label{sec:Data}
The capabilities of GUI agents are fundamentally constrained by data scale, diversity, and quality. To address this, we present AutoTraj, an automated pipeline that generates high-quality GUI interaction trajectories with explicit disturbance awareness. AutoTraj comprises three core components as follows:

\begin{figure*}[t]
\centering
\includegraphics[width=.95\textwidth]{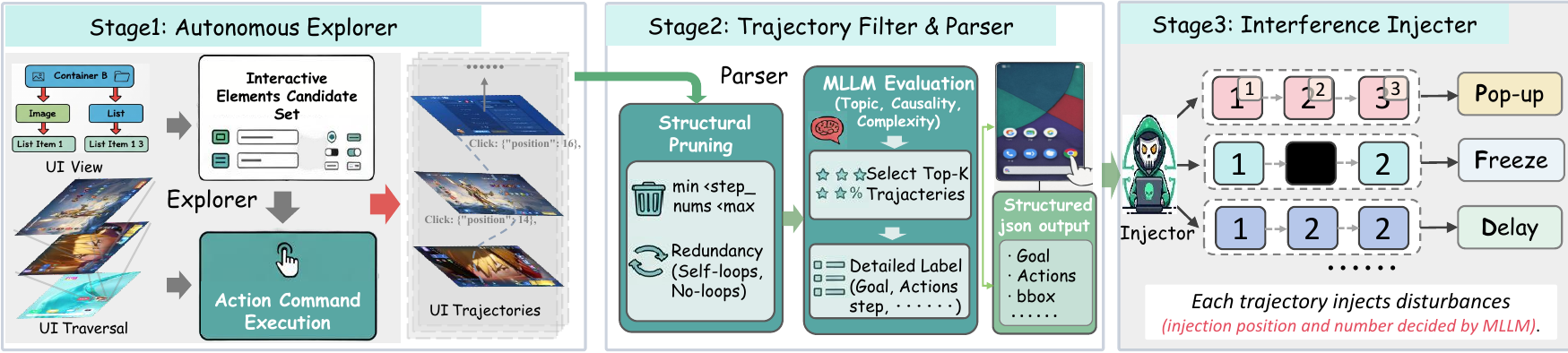} 
\caption{Data Construction. Stage 1 discovers clickable elements via view hierarchy and vision, executes basic actions, and logs replayable UI trajectories. Stage 2 prunes redundant steps, ranks trajectories with an MLLM, and outputs structured annotations. Stage 3 injects realistic disturbances to create clean--perturbed pairs for robustness evaluation.}
\label{fig:data}
\end{figure*}

\mypara{Autonomous Explorer.}
The Explorer efficiently discovers diverse, high-value interaction trajectories using a hybrid perception strategy: it prefers the UI view hierarchy to find actionable elements; when structured signals are missing or incomplete, it falls back to a vision pipeline that combines object detection and optical character recognition (OCR), producing a unified candidate set of controls.

Exploration is driven by a pre-trained agent \cite{ye2025mobileagentv3fundamentalagentsgui} that tries atomic actions (click, scroll) and logs pre- and post-screenshots, as well as action metadata, to produce replayable trajectories. To guide informative exploration, we define the exploration value at state $s_t$ as:

\begin{equation}
V(s_t) = \alpha \cdot \frac{\bigl|E_t \setminus \bigl(\bigcup_{i<t} E_i\bigr)\bigr|}
{|E_t| + \varepsilon}+ (1-\alpha) \cdot \frac{1}{\sqrt{n(s_t)+1}},
\end{equation}
where $E_t$ denotes the control set at $s_t$, $\bigcup_{i<t}E_i$ is the union of controls seen so far, and $n(s_t)$ counts visits to $s_t$. The first term promotes the discovery of unseen controls/layouts, while the second enforces novelty to favor coverage and rarely visited states. $\alpha\in[0,1]$ balances layout discovery and rare-state exploration; hyperparameter analysis appears in Appendix~\ref{sec:exploration_value}.

\mypara{Trajectory Parser.}
To ensure semantic and structural quality, the raw trajectories undergo two-stage filtering and parsing. Stage-1 removes excessively long or redundant trajectories using self-loop and no-op statistics derived from layout changes; their definitions, pruning criteria, and threshold analyses are provided in Appendix~\ref{sec:parser_thresholds}.

\begin{figure*}[t]
\centering
\includegraphics[width=.95\textwidth]{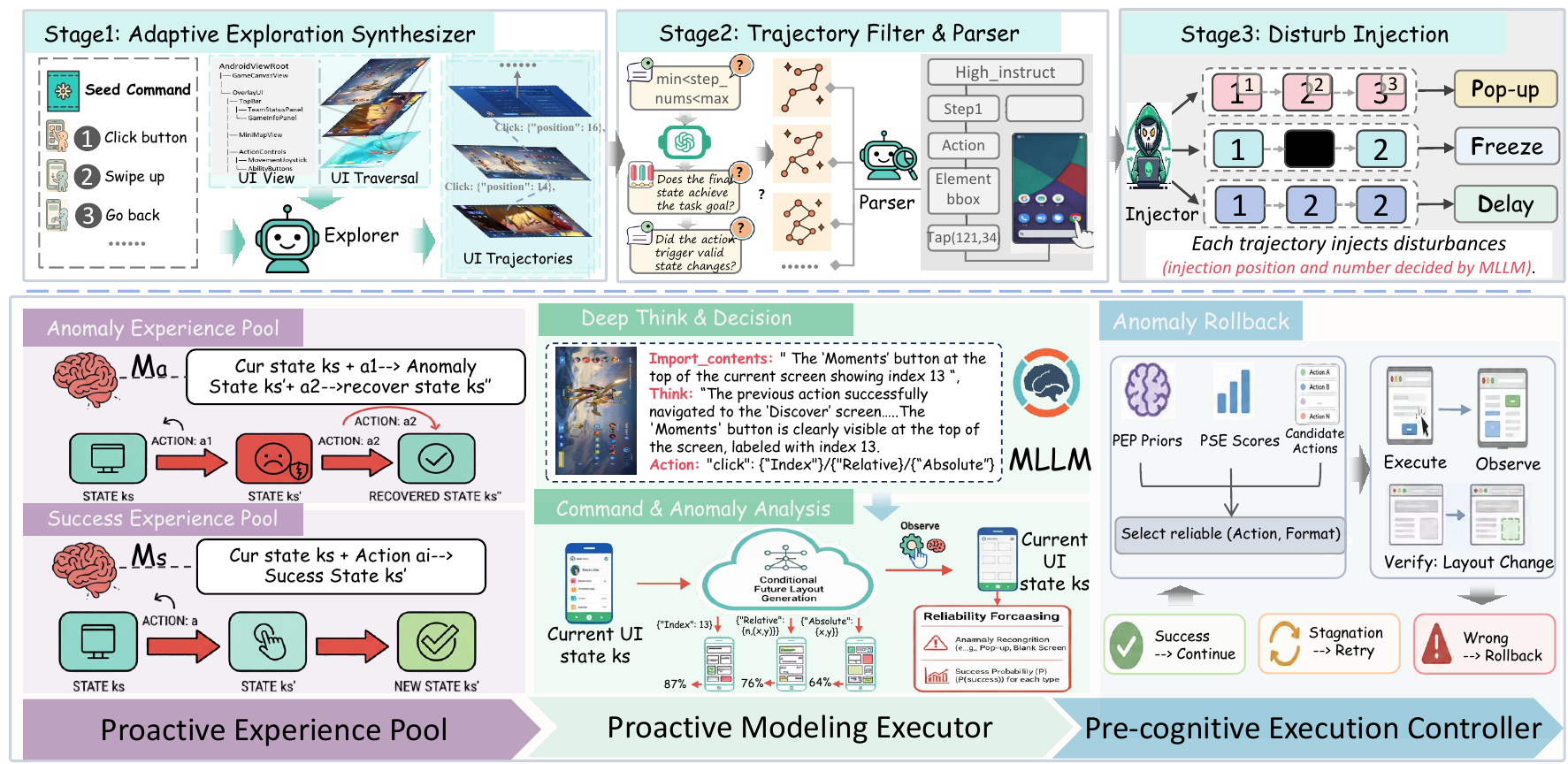} 
\caption{Illustration of PrecogUI. (a) PEP builds a dual memory of anomaly/success patterns; (b) PSE forecasts the next symbolic UI layout and estimates anomaly risk for candidate actions; (c) PEC fuses priors and predictions to prioritize exception handling, select the highest-utility action, and enforce closed-loop monitoring with rollback/retry.}
\label{fig:overview}
\end{figure*}

In Stage-2, we filter trajectories using a high-capacity MLLM \cite{comanici2025gemini} that evaluates topic consistency, causal soundness, and task complexity, selecting the top-$K$ trajectories for detailed labeling.
For each trajectory, the parser yields a high-level goal and stepwise descriptions, exporting structured JSON with goals, step descriptions, action types (normalized coordinates), UI boxes, screen deltas, and execution outcomes for training and evaluation.

\mypara{Perturbation Injector.}
To study robustness, we develop a Perturbation Injector that creates paired samples for evaluation. For each clean trajectory, we randomly inject real-world perturbations covering:
(1) overlay interference (simulating system notifications, pop-up dialogues, etc.);
(2) environmental perturbations (black or repeated frames to simulate loading/lag, and spontaneous layout changes).
All perturbations are screened by six experts to ensure correctness. This yields paired samples for each trajectory: a clean baseline and perturbed variants, enabling comparative evaluation in both ``normal'' and ``perturbed'' modes.

Following this pipeline, AutoTraj produces InterfereBench with 1{,}160 task-level trajectory groups across 34 applications. Each group is anchored by a clean source trajectory of 14--37 steps and includes two controlled perturbation replays of different types. The ${\sim}$27k figure counts annotated screenshots in the clean source trajectories; replayed perturbation frames are generated during evaluation and are not counted again. We use \emph{task group} for the shared instruction and source trajectory, and \emph{execution variant} for its clean or perturbed replay.

To faithfully capture and replay complex interactions beyond single-tap actions, we additionally employ \textbf{PolyTouch} (Appendix~\ref{sec:PolyTouch}), a multi-gesture and macro execution layer that synthesizes deterministic multi-pointer gestures (\eg, three-finger chords, pinch/zoom/rotation) and declarative macros with explicit timing, guards, retries, and rollback.

\subsection{Proactive Experience Pool}
\label{sec:Pool}

We observe that failure-inducing anomaly patterns (\eg, permission pop-ups, network delays) and success-inducing patterns (\eg, app navigation) repeat widely across tasks and applications. Therefore, we propose the Proactive Experience Pool (PEP), which converts costly trial-and-error into efficient experience retrieval. By caching and indexing critical state--action--outcome patterns, the executor can leverage priors rather than plan in isolation. PEP maintains two parallel memories:

(1) \textit{Anomaly Memory ($M_a$):}
$M_a$ records two classes of failures: (i) state--action mappings $(s, a)\mapsto\ell_{\text{anom}}$ when an action in a state yields a specific anomaly; (ii) $s\mapsto\ell_{\text{anom}}$ for states that inherently denote failure (\eg, network outage). Each anomaly entry additionally stores a human-designed or previously successful handling action $a_{\text{handle}}$, which can be reused as a remedy.

(2) \textit{Success Memory ($M_s$):}
$M_s$ stores high-confidence successful transitions $(s,a)\mapsto s'$, indicating that action $a$ in state $s$ reliably reaches a successful successor state $s'$.

\mypara{State Representation and Retrieval Mechanism.}
To enable robust and efficient retrieval, we represent each UI state $s$ by its layout signature: a structured set of interactive elements, with each abstracted as \textit{(type, bbox)}. Instead of relying on exact matches, retrieval is performed by finding the nearest neighbors in the experience pool. Specifically, for a query state $s_t$, we compute the similarity between its layout $L_t$ and each memory layout $L_m$ via a greedy matching algorithm based on element IoU:
\begin{equation}
S_{\text{layout}}(L_t, L_m) = \frac{2|\mathcal{M}|}{|L_t| + |L_m|},
\label{eq:layout-sim}
\end{equation}
where $\mathcal{M}$ is the set of matched pairs identified by the greedy algorithm, which pairs elements from $L_t$ and $L_m$ with the highest Intersection over Union (IoU) score above a predefined threshold. The top-$k$ most similar historical cases are then retrieved to inform the agent. Similarly, actions $a$ are canonicalized based on their type and the target element's normalized coordinates. Crucially, PEP is a dynamic, online-updated memory. Upon encountering new anomalies or discovering successful cases, the agent extracts the state-action-result tuple and asynchronously appends it to the memory pool. We further address cold-start (pool initially empty) and entry staleness (app updates invalidate cached layouts) via a version-aware update policy with exponential staleness decay; empirical analysis is provided in Sec.~\ref{sec:pep_analysis_main}.

\subsection{Proactive Simulation Executor}
\label{sec:PSE}

Mainstream GUI agents are fundamentally reactive and lack foresight into post-action effects. With abrupt transitions in dynamic UIs, reactive policies without anticipation of anomalies tend to fall into irrecoverable failures. Accordingly, we propose the Proactive Simulation Executor (PSE), which forecasts the next symbolic UI layout before acting and evaluates anomaly risk and candidate-action reliability, shifting the paradigm from ``observe--act'' to ``observe--predict--act.''

\mypara{Conditional Future Layout Generation.}

Following ViMo's action-conditioned GUI world-modeling paradigm~\cite{luo2025vimo}, we model one-step UI transitions with a conditional latent diffusion model~\cite{Rombach_2022_CVPR}, fine-tuned on InterfereBench and public GUI datasets~\cite{NEURIPS2024_androidcontrol, 2024_gui_odyssey}. Let $c_{t,j}=\operatorname{Encode}(a_t,r_j)$ denote the concrete command obtained by instantiating candidate action $a_t$ with format $r_j$. PSE predicts $P(\hat{L}_{t+1}^{\,c_{t,j}}\mid L_t,c_{t,j})$. Unlike ViMo's full-screen prediction, PSE generates only an ordered set of \texttt{(type, bbox)} elements for efficient geometric risk tests rather than photorealistic GUI simulation. Architecture, training, and sampling details are provided in Appendix~\ref{sec:diffusion_details}.

\mypara{Reliability Forecasting.} Subsequently, we apply a set of efficient rules to the predicted layout $\hat{L}_{t+1}$ for anomaly recognition. For example, if a bounding box significantly occludes multiple interactive controls in $L_t$, it is flagged as a pop-up anomaly; if interactive elements are nearly absent, it is flagged as a blank-screen anomaly; if $\hat{L}_{t+1}$ remains largely unchanged from $L_t$, the action is likely ineffective or causes freezing.

Motivated by the navigation-style tasks~\cite{gou2025navigating, liu2025_infiguir1, xu2025agenttrek} in our benchmarks, where successful interactions typically induce non-trivial layout changes while failed or frozen steps exhibit near-zero change, we define layout dissimilarity from Eq.~\eqref{eq:layout-sim} as $\mathcal{D}_{\text{layout}}(L,L')=1-S_{\text{layout}}(L,L')$ and use it as a heuristic progress signal. We modulate this signal with the anomaly severity predicted on $\hat{L}_{t+1}^{\,c_{t,j}}$. Specifically, let $w(\hat{L}_{t+1}^{\,c_{t,j}})\in[0,1]$ denote the anomaly-aware reliability weight, which is set to $1$ for non-severe layouts and $0$ for severe anomalies. For each candidate action $a_t$ and command format $r_j$:
\begin{equation}
s(a_t,r_j)=\mathcal{D}_{\text{layout}}(L_t,\hat{L}_{t+1}^{\,c_{t,j}})
\cdot w(\hat{L}_{t+1}^{\,c_{t,j}}).
\label{eq:score}
\end{equation}
The rule-based estimator converts each PSE forecast into an anomaly label and a relative reliability score for ranking candidate action-format pairs; these heuristic scores are not interpreted as calibrated success probabilities. We independently evaluate PSE's prediction quality on held-out test splits, achieving 89.4\% element-type accuracy, 80.2\% mean bbox IoU, and 83.2\% element-level F1 (details in Appendix~\ref{sec:pse_quality}). Note that Eq.~\eqref{eq:score} can misrank edge cases (\eg, confirm dialogs with minimal layout change); the anomaly filter mitigates risk before execution, while PEC verifies residual cases after execution (Sec.~\ref{sec:PEC}).

\begin{table*}[!t]
\centering
\caption{Performance on InterfereBench under two instruction settings (Low/High), $\mathit{TM}_n$ and $\mathit{SR}_n$ denote type-matching and success rate on the normal (non-perturbed) subset; and $\mathit{TM}_a$ and $\mathit{SR}_a$ denote the corresponding metrics on the perturbed subset.}
\label{tab:lh_robust}
\footnotesize
\resizebox{0.88\textwidth}{!}{%
\begin{tabular}{llccccccccc}
\toprule
\multirow{2}{*}{\textbf{Method}} &
\multicolumn{4}{c}{\textbf{InterfereBench-Low}} &
\multicolumn{4}{c}{\textbf{InterfereBench-High}} &
\multicolumn{2}{c}{\textbf{Average}} \\
\cmidrule(lr){2-5}\cmidrule(lr){6-9}\cmidrule(lr){10-11}
& $\mathit{TM_n}$ & $\mathit{SR_n}$ & $\mathit{TM_a}$ & $\mathit{SR_a}$ &
  $\mathit{TM_n}$ & $\mathit{SR_n}$ & $\mathit{TM_a}$ & $\mathit{SR_a}$ &
  $\mathit{TM_m}$ & $\mathit{SR_m}$ \\
\midrule
GPT-4o              & 76.7 & 22.1 & 69.0 & 9.3  & 73.4 & 3.9 & 65.0 & 1.2 & 71.0 & 9.1 \\
Gemini-2.5-Pro          & 87.1 & 28.4 & 80.0 &12.5 & 83.8 &17.7 & 76.0 & 7.5 & 81.7 &16.5 \\
Qwen-2.5-VL         & 90.6 & 33.7 & 82.5 &18.9 & 71.7 &36.5 & 64.0 &18.0 & 77.2 &26.8 \\
\midrule
OmniParser          & 84.1 & 70.9 & 76.0 & 41.5 {\scriptsize\textcolor[HTML]{B38EBB}{($\downarrow$29.4)}} &
                      70.6 & 27.3 & 63.0 & 12.8 {\scriptsize\textcolor[HTML]{B38EBB}{($\downarrow$14.5)}} &
                      73.4 & 38.1 \\
InfiGUI-R1          & 88.9 & 73.6 & 81.5 & 45.8 {\scriptsize\textcolor[HTML]{B38EBB}{($\downarrow$27.8)}} &
                      77.4 & 37.3 & 68.0 & 19.5 {\scriptsize\textcolor[HTML]{B38EBB}{($\downarrow$17.8)}} &
                      79.0 & 44.0 \\
OS-Atlas            & 88.4 & 72.1 & 80.8 & 43.3 {\scriptsize\textcolor[HTML]{B38EBB}{($\downarrow$28.8)}} &
                      72.8 & 30.4 & 63.5 & 14.7 {\scriptsize\textcolor[HTML]{B38EBB}{($\downarrow$15.7)}} &
                      76.4 & 40.1 \\
AgentCPM-GUI        & 90.0 & 75.7 & 82.1 & 46.0 {\scriptsize\textcolor[HTML]{B38EBB}{($\downarrow$29.7)}} &
                      78.9 & 35.7 & 66.8 & 18.4 {\scriptsize\textcolor[HTML]{B38EBB}{($\downarrow$17.3)}} &
                      79.5 & 44.0 \\
UI-TARS-1.5         & 90.8 & 74.5 & 82.6 & 45.0 {\scriptsize\textcolor[HTML]{B38EBB}{($\downarrow$29.5)}} &
                      76.9 & 36.6 & 66.0 & 19.2 {\scriptsize\textcolor[HTML]{B38EBB}{($\downarrow$17.4)}} &
                      79.1 & 43.8 \\
\midrule
\rowcolor[HTML]{F5F2FB}
PrecogUI (Qwen3-VL) & 91.4 & 76.8 & 88.3 & 63.7 {\scriptsize\textcolor[HTML]{C25759}{($\downarrow$13.1)}} &
                    78.6 & 48.9 & 72.0 & 37.2 {\scriptsize\textcolor[HTML]{C25759}{($\downarrow$11.7)}} &
                    82.6 & 56.7 \\
\rowcolor[HTML]{F5F2FB}
\textbf{PrecogUI} & \textbf{92.9} & \textbf{79.2} & \textbf{90.0} & \textbf{68.4} {\scriptsize\textcolor[HTML]{C25759}{($\downarrow$\textbf{10.8})}} &
                    \textbf{80.0} & \textbf{52.7} & \textbf{74.5} & \textbf{41.6} {\scriptsize\textcolor[HTML]{C25759}{($\downarrow$\textbf{11.1})}} &
                    \textbf{84.4} & \textbf{60.5} \\
\bottomrule
\end{tabular}
}
\end{table*}

\subsection{Pre-cognitive Execution Controller}
\label{sec:PEC}
While PSE offers look-ahead predictions, robustness remains uncertain in the absence of a decision framework that converts them into concrete actions. We designa closed-loop controller that fuses PEP priors, PSE predictions, and execution feedback, converting open-ended trial-and-error into guided, self-correcting policy control.

\mypara{Deep Think \& Decision.}
The cycle begins by generating a set of semantically grounded candidate actions $\mathcal{A}$. We steer a base MLLM's reasoning by prompting it to populate a structured JSON schema. This schema mandates a chain of thought that includes: (i) Historical Validation, verifying the outcome of the previous step; (ii) Content Grounding, ensuring that critical UI elements for the current instruction are present; (iii) Think, a step for rationale articulation and failure attribution analysis; and finally (iv) Action, which outputs a ranked set of candidate actions, each with index, relative, and absolute coordinate formats. Expanding each action over its applicable formats yields the candidate-pair set $\mathcal{C}_t=\{(a,r):a\in\mathcal{A},\,r\in\mathcal{R}(a)\}$; the schema and action formats are detailed in Appendix~\ref{sec:prompt}.

\mypara{Pre-cognitive Execution.}
Prior to execution, PEC performs an anomaly check on the current state $s_t$. It first queries the anomaly memory $M_a$ with the current layout. If a sufficiently similar past case is retrieved, the controller triggers the associated remedy $a_{\text{handle}}$; otherwise, deterministic current-layout rules $\mathcal{R}_{\text{anom}}$ check for visible occlusion, blank-screen, and known anomalous-layout patterns. If these rules flag an anomaly, PEC uses a foundation model with a structured anomaly prompt to synthesize a handling action (Appendix~\ref{sec:Workflow_Exception_Handling}). In Algorithm~\ref{alg:pec}, this observed-state check is denoted by $\text{CurrentState}(s_t;\mathcal{R}_{\text{anom}})$ and is distinct from forecasting a candidate action's next layout.

When the state is judged as normal, PEC requests reliability reports for the candidate pairs in $\mathcal{C}_t$ and selects the highest-scoring action-format pair among candidates not flagged as high-risk by PSE.

\mypara{State Monitoring \& Adaptive Recovery}
After executing the action-format pair $(a^*,r^*)$, PEC captures the new state $s_{t+1}$ and verifies the outcome via both the layout change $\mathcal{D}_{\text{layout}}(L_t, L_{t+1})$ and a semantic validation from its MLLM, as part of the subsequent step's Historical Validation. If the execution is judged a failure, PEC triggers recovery according to the failure type:
\begin{itemize}[leftmargin=*]
    \item \textit{Stagnation}: For minimal layout change ($\mathcal{D}_{\text{layout}}(L_t, L_{t+1}) < \tau_c$), PEC treats the command as invalid and retries the action using PSE's next-best command format.
    \item \textit{Unexpected Transition}: If the layout changes significantly ($\mathcal{D}_{\text{layout}}(L_t, L_{t+1}) \geq \tau_c$) but the MLLM's semantic validation deems the new state an incorrect outcome, PEC performs a rollback and adds the failed pair to a temporary taboo list $\mathcal{F}_t$ to block immediate reuse.
\end{itemize}

Overall, PEC selects reliable actions in normal settings and adapts to execution failures. The complete pseudocode is given in Algorithm~\ref{alg:pec} (Appendix~\ref{sec:pec_algorithm}), and a detailed analysis of rollback scope, usage statistics, and handling of irreversible operations is provided in Appendix~\ref{sec:rollback_analysis}.

\section{Experiments}
\label{sec:experiments}

\subsection{Implementation Details}
\label{sec:Details}
We implement PrecogUI on a smartphone UI-automation stack, using Gemini-2.5-Pro as the reasoning backend \cite{comanici2025gemini}. The system is training-free at the agent level: all modules are non-learned except the PSE's next-step layout predictor, which is lightly fine-tuned on InterfereBench and public datasets to model action-conditioned UI transitions. Implementation details are in Appendix~\ref{sec:impl}.

\begin{table*}[!t]
\centering
\caption{Results on AndroidControl and GUI-Odyssey. TM and SR denote type matching and success rate, respectively.}
\label{tab:Navigation_performance}
\footnotesize
\resizebox{0.88\textwidth}{!}{%
\begin{tabular}{lcccccccc}
\toprule
\multirow{2}{*}{\textbf{Method}} &
\multicolumn{2}{c}{\textbf{AndroidControl-Low}} &
\multicolumn{2}{c}{\textbf{AndroidControl-High}} &
\multicolumn{2}{c}{\textbf{GUI-Odyssey}} &
\multicolumn{2}{c}{\textbf{Average}} \\
\cmidrule(lr){2-3} \cmidrule(lr){4-5} \cmidrule(l){6-7} \cmidrule(l){8-9}
& $\mathit{TM}$ & $\mathit{SR}$ & $\mathit{TM}$ & $\mathit{SR}$ &
  $\mathit{TM}$ & $\mathit{SR}$ & $\mathit{TM_m}$ & $\mathit{SR_m}$ \\
\midrule
GPT-4o~\cite{openai2024gpt4ocard} & 74.3 & 19.4 & 63.1 & 21.2 & 37.5 &  5.4 & 58.3 & 15.3 \\
Qwen-2.5-VL~\cite{bai2025qwen25vltechnicalreport} & 94.1 & 85.0 & 75.1 & 62.9 & 59.5 & 46.3 & 76.2 & 64.7 \\
UI-TARS-7B~\cite{qin2025_uitars} & \textbf{98.0} & 90.8 & 83.7 & 72.5 & \textbf{94.6} & 87.0 & \textbf{92.1} & 83.4 \\
\midrule
SeeClick         & 93.0 & 75.0 & 82.9 & 59.1 & 71.0 & 53.9 & 82.3 & 62.7 \\
OS-Atlas-4B~\cite{DOS-ATLAS} & 91.9 & 80.6 & 84.7 & 67.5 & 83.5 & 56.4 & 86.7 & 68.2 \\
InfiGUI-R1~\cite{liu2025_infiguir1} & 96.0 & \textbf{92.1} & 82.7 & 71.1 & --   & --   & 89.4 & 81.6 \\
AgentCPM-GUI~\cite{zhang2025_agentcpm} & 94.4 & 90.2 & 77.7 & 69.2 & 90.9 & 75.0 & 87.7 & 78.1 \\
\midrule
\rowcolor[HTML]{F5F2FB}
\textbf{PrecogUI} &
94.9 & 88.7 & \textbf{86.8} & \textbf{76.4} &
91.3 & \textbf{89.1} & 91.0 & \textbf{84.7} \\
\bottomrule
\end{tabular}
}
\end{table*}

\begin{table}[!t]
\centering
\caption{Grounding performance on ScreenSpot. ``--'' denotes unavailable per-subset results.}
\label{tab:ground_performance}
\footnotesize
\setlength{\tabcolsep}{4pt}
\renewcommand{\arraystretch}{1.05}

\resizebox{\columnwidth}{!}{%
\begin{tabular}{lccccccc}
\toprule
\multirow{2}{*}{\textbf{Method}} &
\multicolumn{2}{c}{\textbf{Mobile}} &
\multicolumn{2}{c}{\textbf{Desktop}} &
\multicolumn{2}{c}{\textbf{Web}} &
\textbf{Avg} \\
\cmidrule(lr){2-3}
\cmidrule(lr){4-5}
\cmidrule(lr){6-7}
& \textit{Text} & \textit{Icon}
& \textit{Text} & \textit{Icon}
& \textit{Text} & \textit{Icon}
& \\
\midrule
GPT-4o      & 30.5 & 23.2 & 20.6 & 19.4 & 11.1 & 7.8  & 18.8 \\
Gemini-2.0  & --   & --   & --   & --   & --   & --   & 84.0 \\
Qwen-2.5-VL & --   & --   & --   & --   & --   & --   & 84.7 \\
\midrule
SeeClick    & 78.0 & 52.0 & 72.5 & 30.0 & 55.7 & 32.5 & 53.4 \\
ShowUI      & 92.3 & 75.5 & 76.3 & 61.1 & 81.7 & 63.6 & 75.1 \\
OmniParser  & 93.9 & 57.0 & 91.3 & 63.6 & 81.3 & 51.0 & 73.0 \\
UI-TARS-7B  & 93.0 & 75.5 & 90.7 & 68.6 & 84.3 & 74.8 & 82.3 \\
InfiGUI-R1  & \textbf{97.1} & 81.2 & 94.3 & 77.1 & 91.7 & 77.6 & 87.5 \\
\midrule
\rowcolor[HTML]{F5F2FB}
\textbf{PrecogUI}
& 96.5
& \textbf{87.8}
& \textbf{97.5}
& \textbf{82.2}
& \textbf{94.6}
& \textbf{91.7}
& \textbf{91.2} \\
\bottomrule
\end{tabular}%
}
\end{table}

\subsection{Benchmarks}
\label{sec:Benchmark}
To evaluate model performance, we use two types of datasets: (1) Self-constructed InterfereBench, designed to test agent robustness in long-horizon and disturbance tasks. It includes two settings: (i) normal, a clean environment; (ii) perturbed, with dynamic perturbations like overlays and layout changes. (2) Public benchmarks, split into (a) ScreenSpot \cite{Cheng2024SeeClick}, which measures UI element grounding (text/icon) to assess localization capability, and (b) navigation-centric suites AndroidControl \cite{NEURIPS2024_androidcontrol} and GUI-Odyssey \cite{2024_gui_odyssey}, which evaluate end-to-end task completion and generalization. AndroidControl provides Low- and High-level instruction settings, whereas GUI-Odyssey uses a single cross-app navigation setting. Evaluation follows standard GUI agent metrics: success rate (SR) and type matching (TM).
See Appendix~\ref{sec:Benchmarks} for benchmark details.

\subsection{Main Results}
\label{sec:results}

\mypara{Perturbation Handling.} We evaluate PrecogUI on InterfereBench to assess long-horizon and dynamic-UI robustness. We compare against: 1) base models, GPT-4o \cite{openai2024gpt4ocard}, Gemini-2.5-Pro \cite{comanici2025gemini}, and Qwen-2.5-VL \cite{bai2025qwen25vltechnicalreport}; and 2) specialized GUI agents, including OmniParser \cite{Wan_2024_CVPR}, InfiGUI-R1 \cite{liu2025_infiguir1}, AgentCPM-GUI \cite{zhang2025_agentcpm}, OS-Atlas \cite{DOS-ATLAS}, and UI-TARS \cite{qin2025_uitars}. As shown in Table~\ref{tab:lh_robust}, on the normal subset, PrecogUI reaches an SR of 79.2\% (Low) and 52.7\% (High), outperforming the best reactive GUI agent by 3.5 and 15.4 percentage points, respectively. On the perturbed subset, PrecogUI degrades by only 10.8 and 11.1 points (Low/High) and outperforms the strongest baseline by 22.4 and 22.1 points, respectively.

\mypara{Grounding Capability.} Table~\ref{tab:ground_performance} reports ScreenSpot grounding accuracy across mobile, desktop, and web interfaces. PrecogUI achieves the best sample-weighted average accuracy of 91.2\% and leads five of the six displayed subsets; InfiGUI-R1 remains 0.6 points higher on Mobile Text. Overall, PrecogUI exceeds InfiGUI-R1's weighted average by 3.7 points. Further discuss in Appendix~\ref{sec:cross_device}.

\mypara{Navigation Capability.} To validate generalization, PrecogUI is evaluated on AndroidControl and GUI-Odyssey. As shown in Table~\ref{tab:Navigation_performance}, PrecogUI achieves SRs of 88.7\% (Low) and 76.4\% (High) on AndroidControl and 89.1\% on GUI-Odyssey. Relative to Qwen-2.5-VL, the gains on AndroidControl-Low and -High are 3.7\% and 13.5\%. Compared with UI-TARS-7B, PrecogUI is 2.1 points lower on AndroidControl-Low but 3.9 and 2.1 points higher on AndroidControl-High and GUI-Odyssey, yielding a 1.3-point gain in average SR over the three datasets. Moreover, Table~\ref{tab:lh_robust} shows that replacing the reasoning backbone with Qwen3-VL~\cite{bai2025qwen3vl} yields slightly lower yet competitive results, consistently surpassing the specialized GUI-agent baselines on InterfereBench and indicating that the robustness gains are not tied to a single reasoning backend.

\begin{figure*}[!t]
\centering
\includegraphics[width=.95\textwidth]{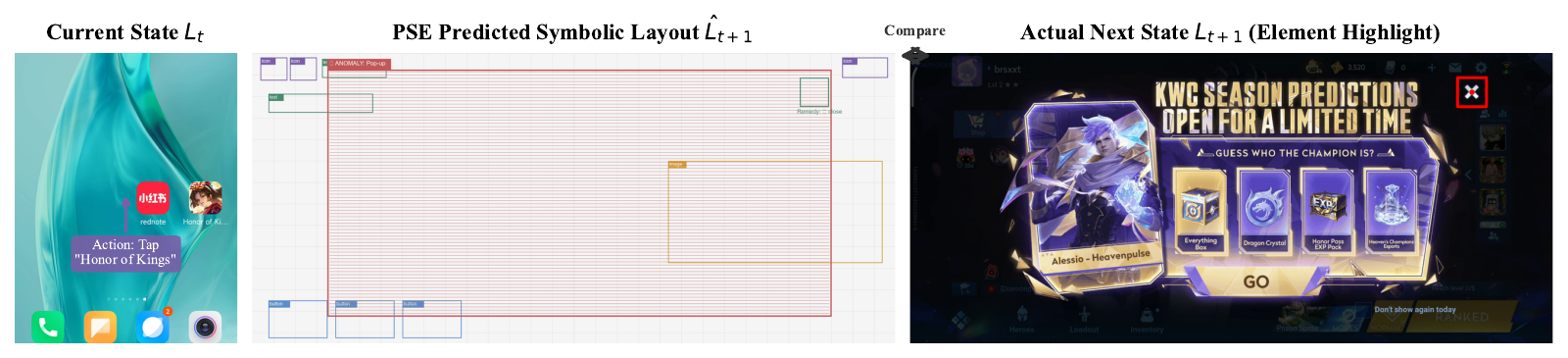}
\caption{PSE prediction quality. Left: current state $L_t$; center: predicted symbolic layout $\hat{L}_{t+1}$; right: actual next state.}
\label{fig:pse_quality}
\end{figure*}

\begin{figure*}[!t]
\centering
\includegraphics[width=.95\textwidth]{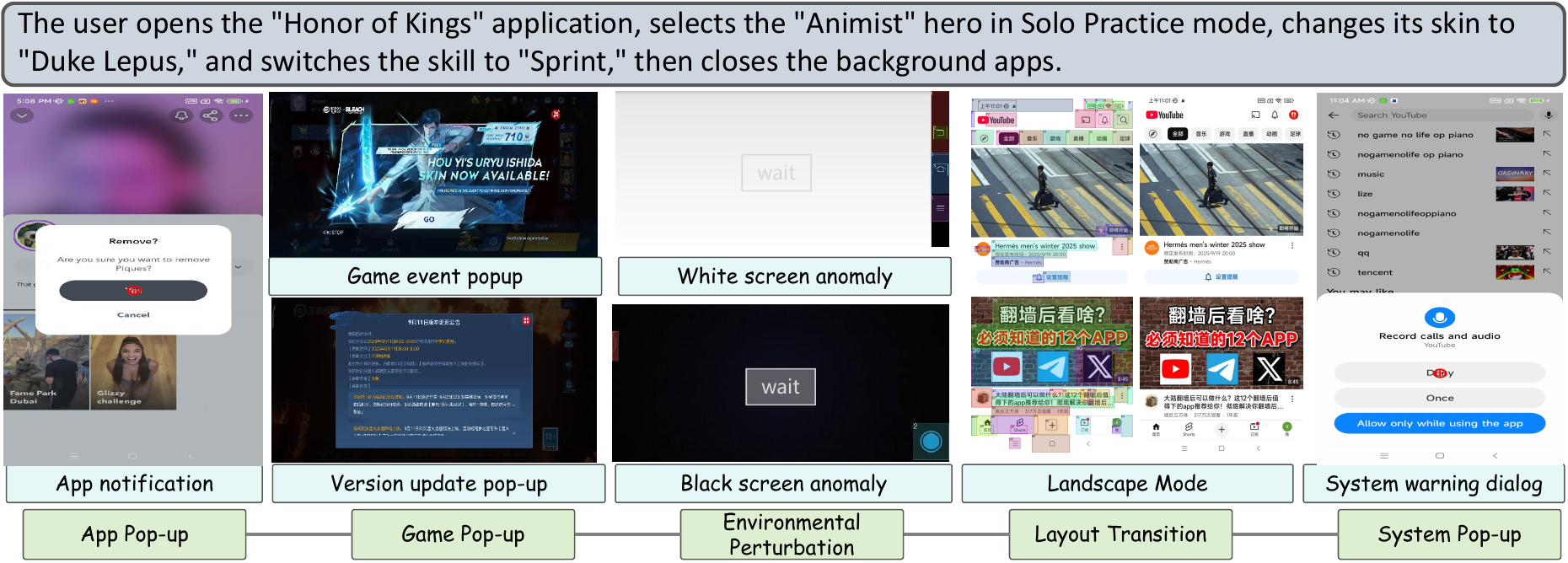}
\caption{Case visualization of PrecogUI under representative anomalous scenarios.}
\label{fig:visualization}
\end{figure*}

\subsection{Ablation Study}
\label{sec:Ablation}

\mypara{Effectiveness of the Experience Pool (PEP) and Incremental Components.}
To assess the contributions of PrecogUI components, we conduct cumulative ablations on the InterfereBench strong-perturbation subset. As shown in Figure~\ref{fig:ablation}(a), the reactive baseline achieves 41.5\% SR; adding PEP, PSE, and PEC successively raises SR to 52.1\%, 61.8\%, and 68.4\%, respectively. The corresponding incremental gains are 10.6, 9.7, and 6.6 percentage points. On the clean long-horizon subset, SR similarly increases from 70.9\% to 74.8\%, 77.1\%, and 79.2\%. These controlled results show the incremental contribution of next-layout forecasting within the complete pipeline. Further reliability and PEP analyses are provided in Appendix~\ref{sec:PSE Analysis}.


\mypara{PSE Prediction Quality.}
\label{sec:pse_quality_main}
We evaluate PSE's predicted layouts against ground truth on held-out test splits. Figure~\ref{fig:pse_quality} shows a representative case: given the current state $L_t$ (left) and a tap action, PSE generates a \emph{symbolic} layout $\hat{L}_{t+1}$ (center), containing only \texttt{(type, bbox)} elements on a blank canvas. The rule-based estimator then flags the predicted occlusion as a pop-up anomaly, while PEC supplies the handling action, consistent with the actual next state (right).

\mypara{Case Visualization.}
\label{sec:Visualization}
As illustrated in Figure~\ref{fig:visualization}, PrecogUI robustly handles anomalies across cross-application, strongly perturbed, and dynamic GUI tasks, including application and in-game pop-ups, black or white screens, and abrupt layout shifts caused by system-level interruptions. Beyond reacting to visible disturbances, it predicts potential layout changes and latent anomalies. During loading delays, for example, it waits rather than interacting with blank or unresponsive screens. This predictive avoidance enables reliable execution in complex workflows, distinguishing PrecogUI from conventional reactive agents.
More cases are provided in Appendix~\ref{sec:Qualitative}.

\section{Related Work}
\label{sec:related_work_main}
Multimodal large language models (MLLMs)~\cite{Li2023BLIP2, Liu_2024_CVPR, Chen_2024_CVPR} provide the visual-language representations used by many GUI agents, yet models trained primarily on static perception tasks do not by themselves provide persistent state tracking in dynamic interfaces.
Building upon MLLMs, GUI agents~\cite{gou2025navigating, liu2025_infiguir1, Lin_2025_CVPR, kang2026longhorizonui} learn sequential action policies by mapping instruction--screenshot pairs to grounding and interaction. Recent efforts further introduce structure and memory: AutoDroid~\cite{AutoDroid_AICMCN_2025} injects app-specific knowledge collected through exploration, while MapAgent~\cite{MapAgent_kong_2025} retrieves structured page memories during planning. ViMo~\cite{luo2025vimo} instead predicts full future GUI observations for candidate-action selection.
PrecogUI follows this proactive world-modeling direction but targets perturbation-robust execution: its lightweight symbolic predictor supplies geometric risk cues, while PEP and PEC provide experience retrieval and closed-loop recovery. A comprehensive discussion is provided in Appendix~\ref{sec:Related_work}.

\section{Concluding Remarks}
In this work, we present PrecogUI for reliable and adaptive long-horizon GUI execution under frequent and diverse disturbances. Through an experience--foresight--feedback loop, PEP retrieves prior success and anomaly patterns, PSE predicts future layouts and ranks candidate actions by estimated reliability, and PEC enables monitored execution with retry and rollback. Experiments on InterfereBench and public benchmarks demonstrate improved task success and robustness in complex, dynamic environments.

\bibliography{main}

\newpage
\appendix

\section*{Appendix}

This is the supplementary file for our submission titled \textit{PrecogUI: Proactive GUI agents via Pre-cognitive Simulation and Experience Retrieval}. This material supplements the main paper with the following content:
\vspace{0.3cm}
\begin{itemize}
   \item (\ref{sec:motivation}) \textbf{Motivation of PrecogUI}

    \item (\ref{sec:Related_work}) \textbf{Related work}
    \item (\ref{sec:PolyTouch}) \textbf{PolyTouch: A Multi-Gesture and Macro Execution Layer}

    \item (\ref{sec:Additional Experiments}) \textbf{Additional Experiments}
    \begin{itemize}
        \item (\ref{sec:impl}) Implementation Details
        \item (\ref{sec:Benchmarks}) Benchmarks
        \item (\ref{sec:diffusion_details}) Details of Diffusion-based Future Layout Generation
        \item (\ref{sec:exploration_value}) Hyperparameter analysis of the exploration value
        \item (\ref{sec:parser_thresholds}) Hyperparameter analysis of parser thresholds
        \item (\ref{sec:pse_quality}) PSE prediction quality evaluation
    \end{itemize}
    \item (\ref{sec:prompt}) \textbf{Prompts in Automated Pipeline}
       \begin{itemize}
        \item (\ref{sec:output_Format}) Output Format Structure Template
        \item (\ref{sec:Action_Selection_Template}) Action Selection Template
        \item (\ref{sec:role_context}) Role and Context Template
        \item (\ref{sec:Workflow_Exception_Handling}) Anomaly Handling Template
        \item (\ref{sec:os_hint}) OS-Specific Hints
        \item (\ref{sec:general_instructions}) General Instructions
    \end{itemize}

    \item (\ref{sec:Qualitative}) \textbf{Qualitative Analysis}

    \item (\ref{sec:pec_algorithm}) \textbf{PEC Algorithm Pseudocode}

    \item (\ref{sec:discussion}) \textbf{Additional Discussions}
    \begin{itemize}
        \item (\ref{sec:cross_device}) Cross-Device Grounding and Mobile Navigation Analysis
        \item (\ref{sec:rollback_analysis}) Rollback Scope and Empirical Statistics
    \end{itemize}
\end{itemize}



\section{Motivation of PrecogUI}
\label{sec:motivation}

\begin{figure*}[t]
\centering
\includegraphics[width=1.0\linewidth]{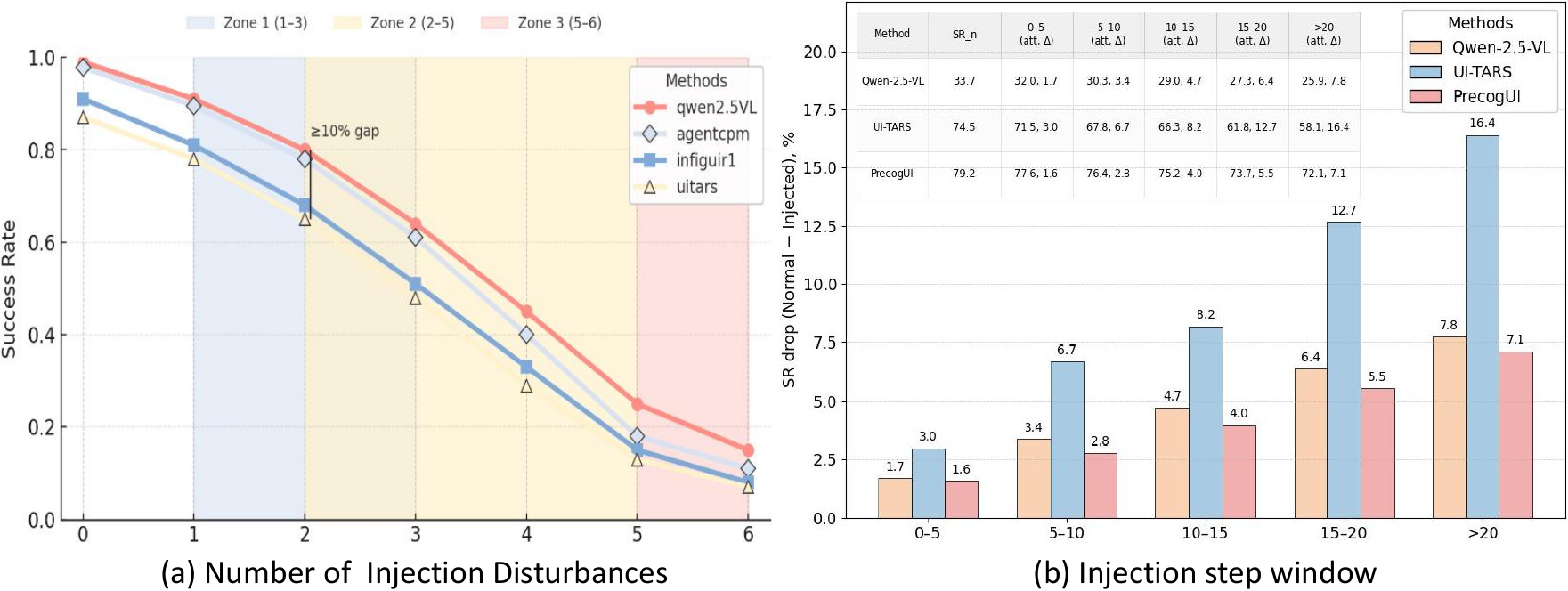} 
\caption{Impact of disturbance count and timing on policy success rates. The left panel shows SR degradation with increasing disturbance count. The right panel illustrates the greater sensitivity of reactive policies to later disturbance injections, in contrast to \textsc{PrecogUI}'s robustness.}
\label{fig:disturbance_sensitivity}
\end{figure*}

Figure~\ref{fig:disturbance_sensitivity} reveals two critical patterns. First, the success rate (SR) declines sharply with an increasing number of injected disturbances. Reactive baselines plummet from nearly 100\% SR to below 20\% with zero to six injections, showing a performance gap of at least 10\% by just two injections (left panel). This highlights the inherent brittleness of purely reactive policies under sustained interference. Second, disturbance timing significantly impacts performance (right panel). Shifting a single injection later in the trajectory yields greater SR losses across all baselines. For instance, \textsc{UI-TARS} exhibits an SR drop escalating from 3.0\% (steps 0--5) to 16.4\% ($>20$ steps). In contrast, \textsc{PrecogUI} demonstrates consistent resilience, increasing only from 1.6\% to 7.1\%---approximately $2.3\times$ less degradation than \textsc{UI-TARS} in the long-horizon tail---while maintaining higher nominal SR. These trends suggest that coupling experience priors with look-ahead simulation is crucial for mitigating late-stage error cascades.

\section{Related Work}
\label{sec:Related_work}

\mypara{Multimodal Large Language Models.} MLLMs \cite{Li2023BLIP2, Liu_2024_CVPR, Chen_2024_CVPR} have emerged as a central enabler for GUI automation, boosting both perceptual and reasoning capabilities of agents. By parsing complex screen structures and grounding natural-language instructions in UI elements, MLLMs serve as the perception backbone for mainstream agents \cite{NEURIPS2024_0520537b, Yang_2025_CVPR}. Benchmarks such as MMMU~\cite{Yue_2024_CVPR} measure broad multimodal understanding, while task-specific studies evaluate capabilities such as visual question answering~\cite{10438044} and image captioning~\cite{Dai2023InstructBLIP}. These static or single-turn capabilities are useful foundations but do not alone provide persistent state tracking or anticipatory control in dynamic interfaces.

\mypara{GUI agents.} Research on GUI agents \cite{gou2025navigating, liu2025_infiguir1, xu2025agenttrek} has explored diverse strategies for policy learning and grounding. A common paradigm~\cite{Lin_2025_CVPR, kang2026longhorizonui} is to fine-tune multimodal models to map instruction--screenshot inputs into sequential action predictions. For example, UGround \cite{gou2025navigating} trains a purely visual grounding model on millions of UI elements, enabling click and operation solely through visual localisation. Recent efforts \cite{Gao_2025_arXiv, Zhang_2025_TOIS} have added structure and memory: AutoDroid~\cite{AutoDroid_AICMCN_2025} injects app-specific knowledge collected through automated exploration, and MapAgent~\cite{MapAgent_kong_2025} retrieves structured page memories during planning. While effective on common GUI benchmarks \cite{gao2024mobileviewslargescalemobilegui}, these methods \cite{lei2025grounding, xu2025aguvis} remain largely reactive, leaving them vulnerable to unforeseen perturbations. An unexpected pop-up can hijack the agent's attention, while a loading delay may be misinterpreted as a failed action.

\mypara{Adjacent Multimodal and Proactive-Agent Research.}
Several studies outside direct GUI action prediction provide complementary perspectives. CalibCLIP~\cite{CalibCLIP_2025} calibrates dominant visual and textual tokens for text-driven image retrieval, illustrating how representation-level calibration can reduce misleading visual semantics, although it does not model GUI transitions. Proactive Agent~\cite{Lu_2025_ICLR} studies when an LLM agent should initiate assistance from contextual signals rather than waiting for an explicit request; its notion of proactivity is conceptually related but differs from PrecogUI's action-conditioned UI forecasting. AgentSteerTTS~\cite{kang2026agentsteerttsmultiagentclosedloopframework} demonstrates a multi-agent closed loop for composite-instruction speech synthesis. We cite it only as a cross-domain example of iterative feedback control, not as a GUI perception or execution baseline.

\mypara{GUI world models.} ViMo~\cite{luo2025vimo} predicts full future GUI observations for action selection. PrecogUI adopts the same action-conditioned foresight principle but predicts lightweight symbolic layouts tailored to geometric anomaly detection and closed-loop recovery.

\begin{figure*}[!ht]
\centering
\includegraphics[width=0.8\linewidth]{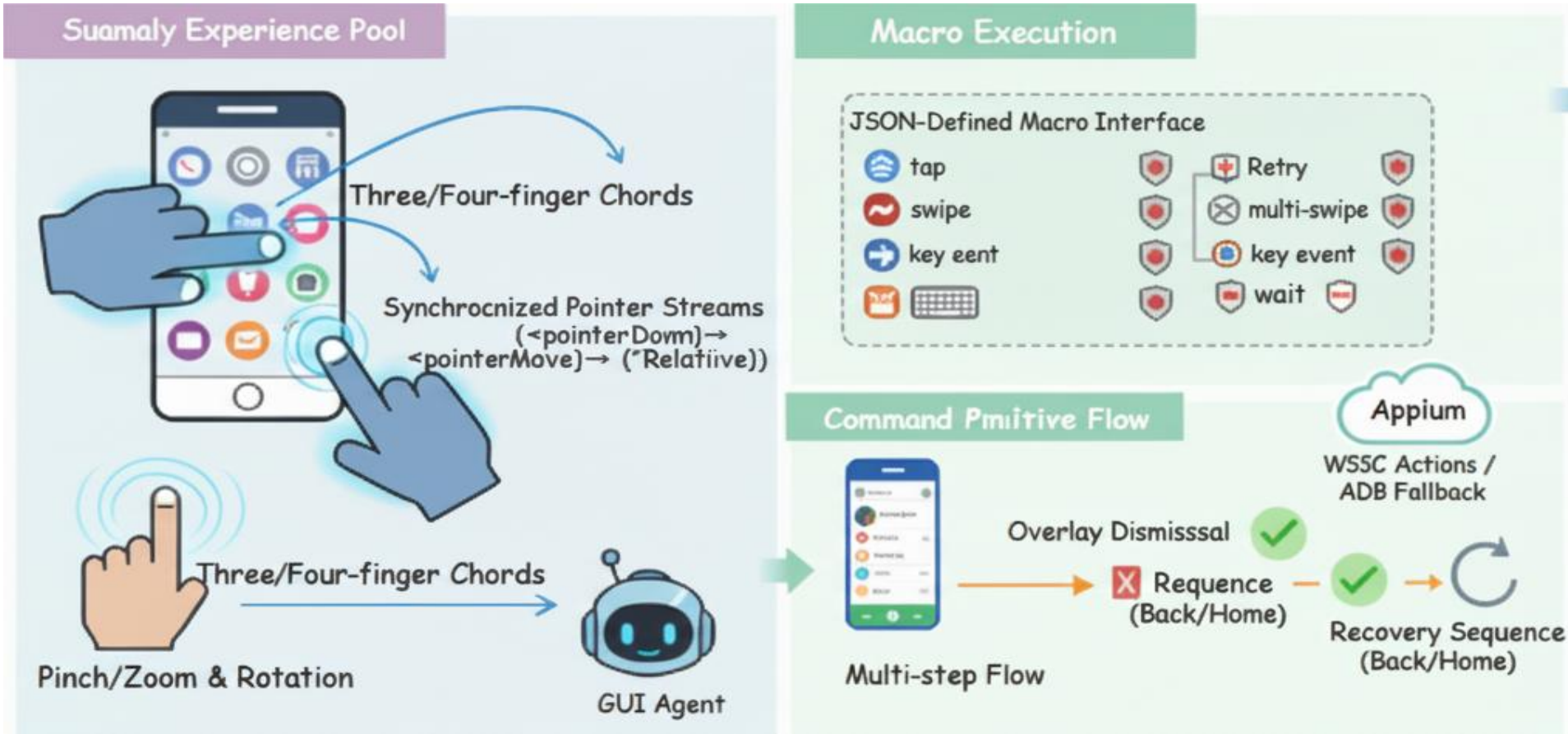} 
\caption{The illustration of PolyTouch, a multi-gesture and macro execution layer for GUI agents. It depicts multi-finger gestures and macro-level commands, highlighting their role in robust, long-horizon task execution.}
\label{fig:PolyTouch}
\end{figure*}

\section{PolyTouch: A Multi-Gesture and Macro Execution Layer}
\label{sec:PolyTouch}

Real-world mobile applications often require multi-pointer and multi-step interactions, such as three- or four-finger system shortcuts, pinch/zoom and rotation in media and map viewers, or coordinated sequences in creative tools. Existing GUI agents generally assume single-touch atomic operations and one-shot execution, which makes them fragile when facing complex gestures, long interaction flows, or OS-level controls that demand precise synchronization. To address this gap, we introduce \textbf{PolyTouch}, an execution layer that extends the action space to multi-finger gestures and macro-level commands with explicit timing, guards, and rollback mechanisms.

PolyTouch supports a wide range of interaction patterns rarely considered in prior work:
(i) Multi-finger chords for dialogs, split-screen, or editing shortcuts;
(ii) Continuous gestures such as pinch, zoom, and rotation;
(iii) Multi-step flows with explicit waiting, retries, and overlay dismissal;
(iv) Recovery sequences (\eg, back, home, or targeted close) that must be executed atomically to exit unexpected states.
These abstractions allow agents to operate robustly in long-horizon tasks where traditional atomic actions fail.

PolyTouch builds on Appium's W3C Actions API for deterministic multi-pointer synthesis and it falls back to ADB when accessibility channels are blocked. Its design centers on:
(1) deterministic timing through tick-based scheduling;
(2) unified coordinate formats (index, relative-in-box, absolute) with boundary-safe mapping;
(3) a declarative macro interface that bundles taps, swipes, multi-swipes, key events, and waits into atomic, retryable units;
(4) graceful degradation to equivalent ADB commands while preserving ordering and timing.

PolyTouch exposes two main capabilities:
\textbf{(a) Multi-gesture execution.} Three- and four-finger gestures are represented as synchronized pointer streams (\texttt{pointerDown} $\rightarrow$ \texttt{pointerMove} $\rightarrow$ \texttt{pointerUp}), while pinch/zoom and rotation are parameterized around target boxes and derived from relative coordinates.
\textbf{(b) Macro execution.} JSON-defined macros encapsulate an ordered list of primitives with explicit guard, retry, and rollback semantics, supporting flexible coordinate specifications.

PolyTouch integrates into the agent control loop by providing reliability-aware plans and structured execution reports (success flags, layout changes, anomaly tags). These outputs feed the Proactive Experience Pool to accumulate reusable patterns and guide the Pre-cognitive Execution Controller in anticipating failures and triggering recovery. In this way, PolyTouch transforms low-level taps into a closed-loop, macro-level control primitive that is both expressive and robust.



\section{Additional Experiments}
\label{sec:Additional Experiments}

\subsection{Implementation Details}
\label{sec:impl}

\mypara{Hardware \& Devices.}
All experiments were conducted on a single training node with \textbf{8$\times$ NVIDIA H20 (96\,GB)} GPUs. For on-device evaluation, we used a pool of mainstream Android phones covering \textbf{Huawei/Honor}, \textbf{Xiaomi/Redmi}, and \textbf{OPPO/realme}, spanning Android~10--14 and common resolutions (720p--1440p). Devices were connected over USB with \texttt{ADB} (USB debugging enabled) for reliable screenshot capture and input dispatch; Wi-Fi \texttt{ADB} was used only for long-duration soak tests.

\mypara{Data Collection \& Real-World Tests.}
We employ \textbf{Appium~2.x} (Android driver: \texttt{uiautomator2}) together with \texttt{ADB} to (i) scrape view hierarchies and screenshots, (ii) execute action sequences in real apps, and (iii) log pre/post frames, timing, and outcomes for replayable trajectories. For latency-critical fallback (\eg, when Appium is blocked by transient overlays), we issue low-level commands via \texttt{adb shell input} (tap/swipe/keyevent) and re-sync with Appium on the next stable frame. Randomized perturbation placement, candidate ordering, and PSE initialization use fixed seeds; device capture settings are held constant. Screen coordinates are normalized to $[0,1]$ and mapped to device pixels at runtime.

\mypara{Evaluation Isolation.}
Each trajectory corpus is partitioned before PSE training, and no transition from a held-out evaluation trajectory is included in the training split. For every reported test episode, PEP is initialized from the same training-split memory and is reset to that state before the episode begins. Updates collected within an episode may guide later steps of that episode but are discarded at termination; therefore, no memory entry created from one test episode can be retrieved in another. The same held-out task split and action budget are used for all methods within each benchmark.

\begin{figure*}[!ht]
\centering
\begin{subfigure}[t]{0.48\linewidth}
    \centering
    \includegraphics[width=\linewidth]{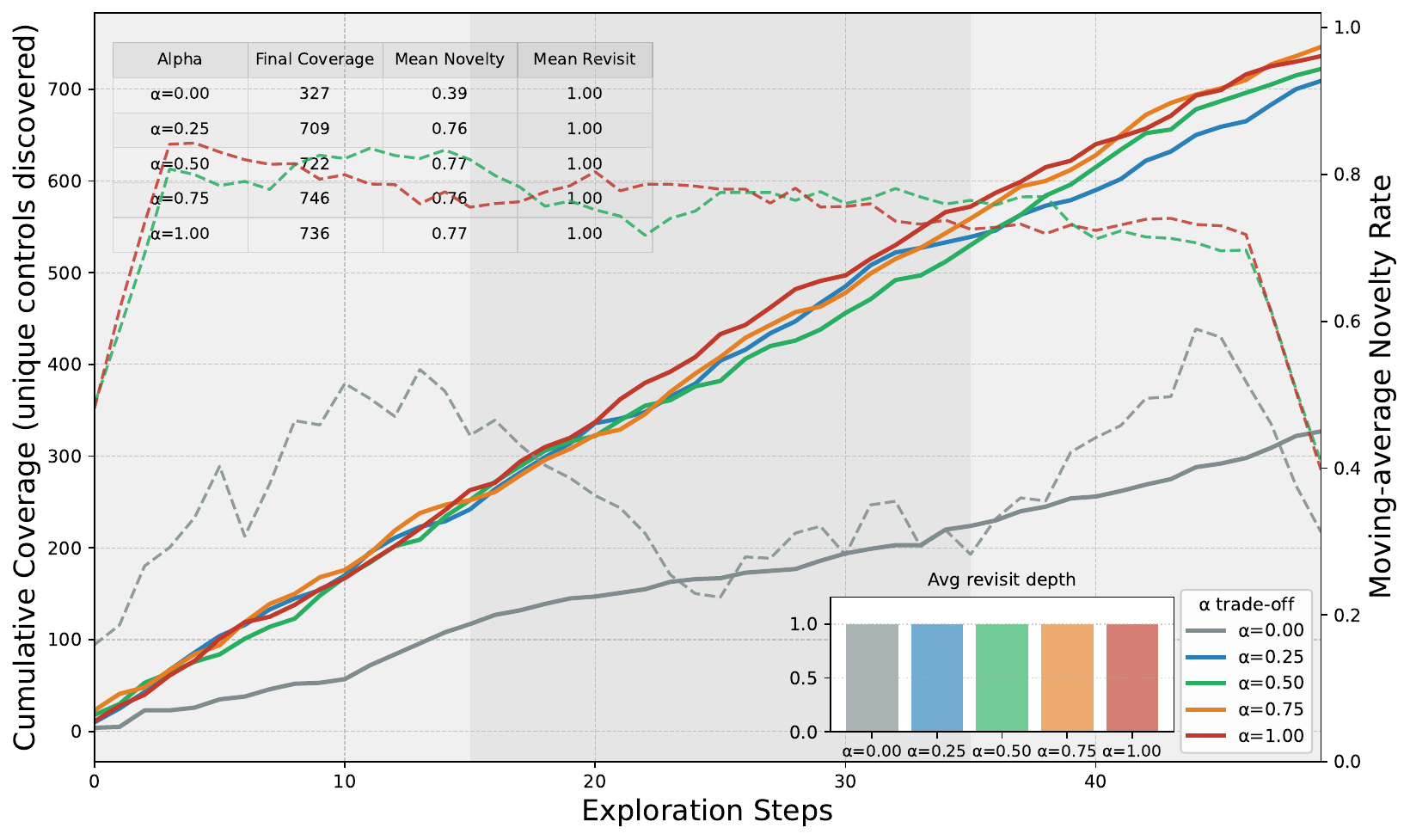}
    \caption{Effect of exploration weight $\alpha$ on 50-step coverage and novelty.}
    \label{fig:alpha_exploration_50steps}
\end{subfigure}
\hfill
\begin{subfigure}[t]{0.48\linewidth}
    \centering
    \includegraphics[width=\linewidth]{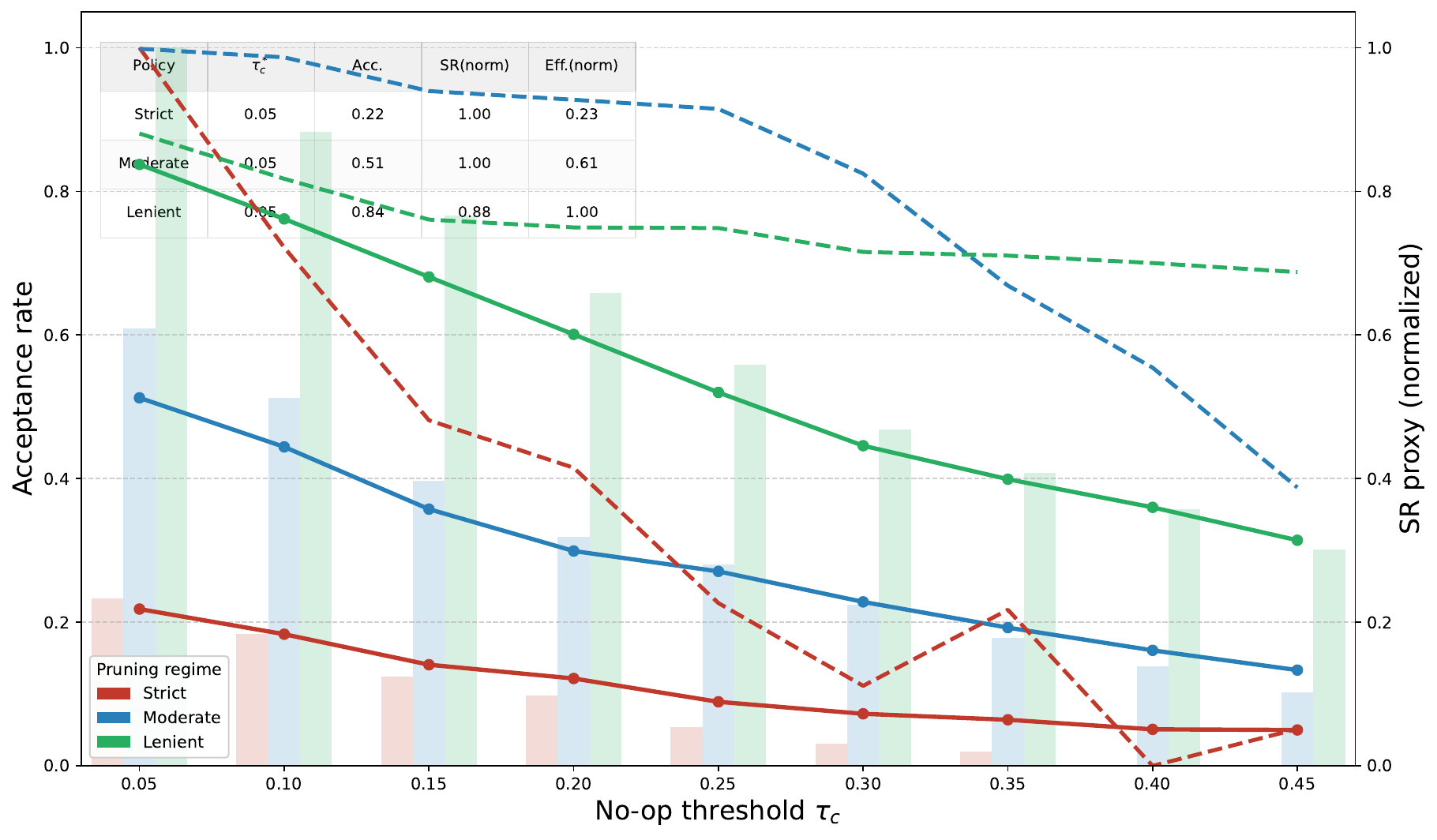}
    \caption{Trade-off of Stage-1 parser acceptance under different $\tau_c$.}
    \label{fig:parser_thresholds_tradeoff}
\end{subfigure}
\caption{Ablation of exploration value and parsing thresholds. (a) Larger $\alpha$ accelerates unique-control coverage but reduces rare-state revisits. (b) Moderate pruning near $\tau_c\!\in[0.25,0.35]$ gives the best acceptance--success--efficiency balance.}
\label{fig:alpha_parser_ablation}
\end{figure*}

\subsection{Benchmarks}
\label{sec:Benchmarks}

\mypara{Grounding-Centric Benchmark: ScreenSpot.}
Accurate element localization is the foundation of GUI automation. ScreenSpot~\cite{Cheng2024SeeClick} is a cross-platform grounding benchmark with over 1,200 natural-language instructions spanning iOS, Android, macOS, Windows, and Web interfaces. Each instruction is paired with pixel-level bounding boxes and element-type labels (text, icon, or widget) and covers challenging scenarios such as icon-text composites and occluded controls.

\mypara{Navigation-Centric Benchmarks: AndroidControl \& GUI Odyssey.}
Once elements can be reliably located, agents must navigate within and across apps. AndroidControl~\cite{NEURIPS2024_androidcontrol} contains 15,283 human demonstrations of everyday Android tasks. Each task is paired with a high-level goal instruction and a low-level, step-by-step instruction; the Low/High labels in our tables refer to instruction granularity rather than separate single-app and cross-app difficulty partitions. We use the GUI-Odyssey v1 protocol~\cite{2024_gui_odyssey}, which contains 7,735 cross-app episodes collected on six mobile devices and spans 201 apps and approximately 1.4K app combinations. GUI-Odyssey is reported as a single cross-app navigation setting and is not assigned AndroidControl-style Low/High labels.

\mypara{Disturbance-Aware Benchmark: InterfereBench.} InterfereBench covers 34 applications---complex games, enterprise tools, and general apps---with bilingual (Zh/En) UIs recorded on diverse phone models. It contains 1{,}160 task-level trajectory groups whose clean source trajectories span 14--37 steps and contain 27{,}124 annotated screenshots. We captured 574 real abnormal screens and curated 217 synthetic disturbance assets (pop-ups, notifications, black screens, and layout shifts). Each task group comprises one clean execution and two controlled perturbation replays; replayed frames are generated online and are not added to the source-screenshot count. Annotations include high-level goals and step-level structures (action type, normalized coordinates, UI boxes, screen deltas, and outcomes).

\begin{figure*}[!t]
\centering
\includegraphics[width=.98\textwidth]{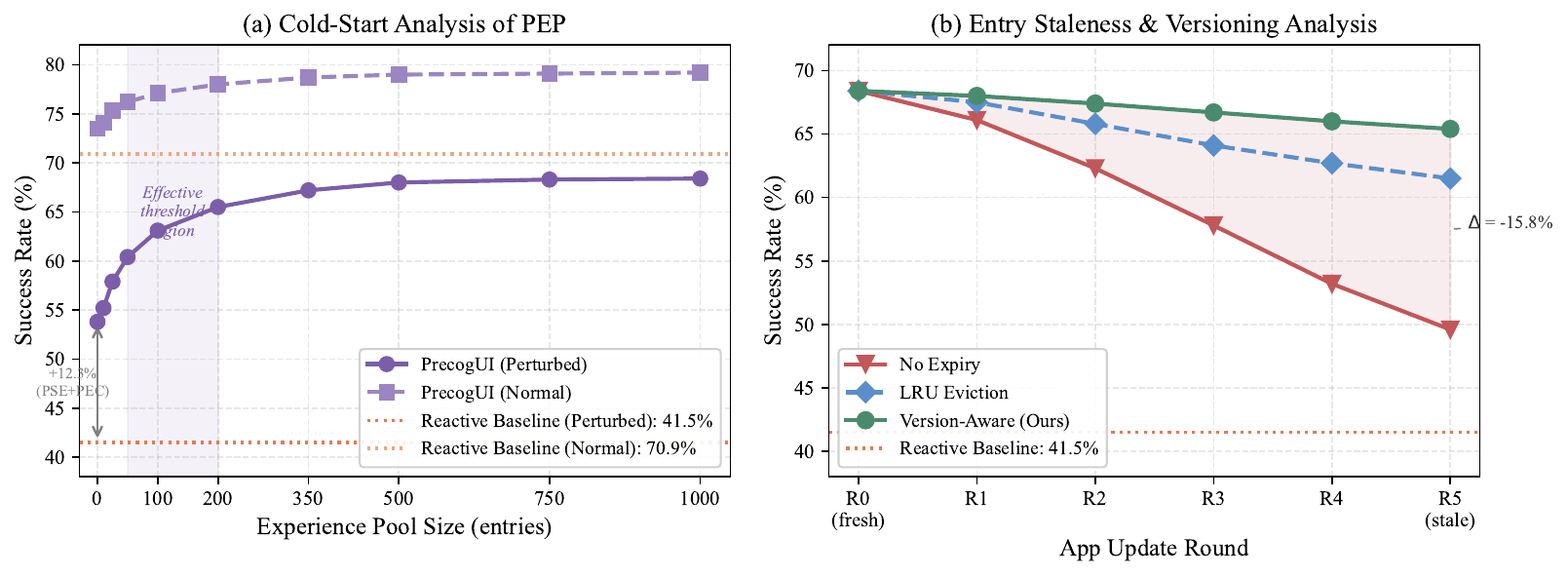}
\caption{In-depth analysis of PEP. (a) SR as a function of experience pool size, showing rapid gains up to 200 entries and saturation beyond 500. (b) Memory-management strategy comparison across app-update rounds; the version-aware policy best preserves effectiveness.}
\label{fig:pep_analysis}
\end{figure*}

\subsection{Further Analysis of Proactive Reliability Mechanisms}
\label{sec:PSE Analysis}

\begin{figure*}[!t]
\centering
\includegraphics[width=.98\textwidth]{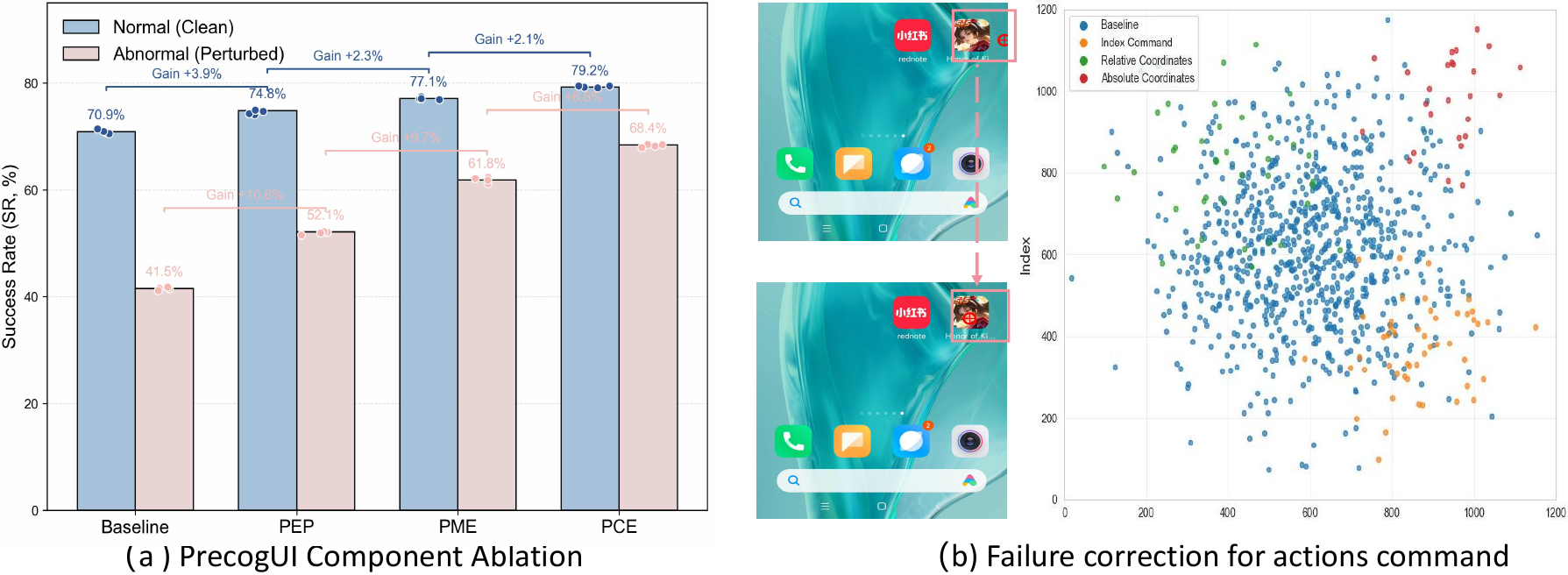}
\caption{Ablation and reliability analysis of PrecogUI. (a) Cumulative component contributions on clean and strong-perturbation subsets; the legacy labels PME and PCE in the exported panel correspond to PSE and PEC, respectively. (b) Representative command-format failure correction using PSE reliability forecasts.}
\label{fig:ablation}
\end{figure*}

\mypara{Analysis of Proactive Reliability Forecasting.}
PSE forecasts the next-step layout for each concrete index, relative, or absolute command and assigns a relative reliability score. Figure~\ref{fig:ablation}(b) illustrates how the resulting risk signal supports command-format correction. Thus, the hierarchy Index $\rightarrow$ Relative-in-Box $\rightarrow$ Absolute is a default preference rather than an immutable order: PSE may override it when another concrete command is predicted to be safer in the current context.

\begin{figure*}[!t]
\centering
\includegraphics[width=0.9\linewidth]{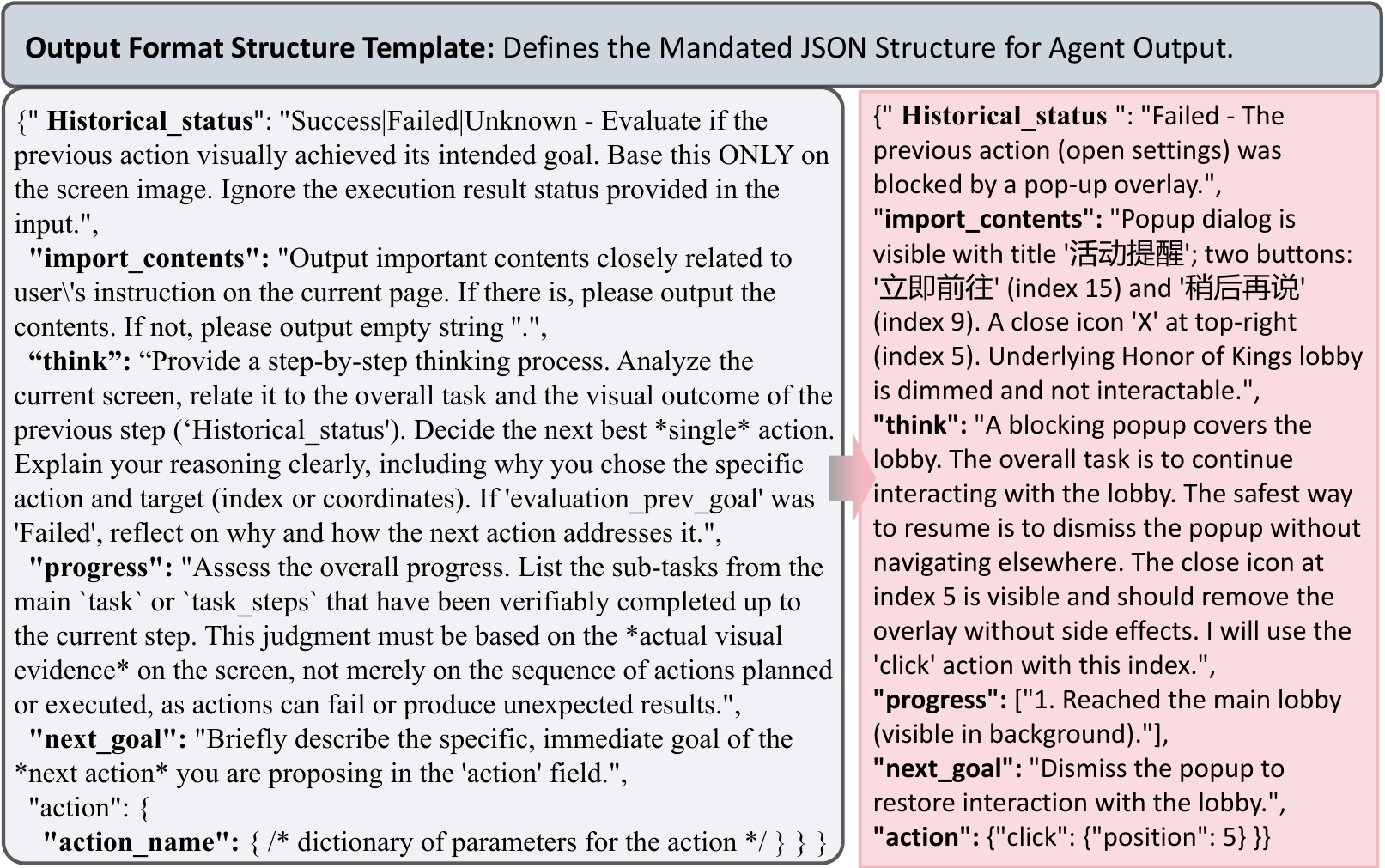} 
\caption{Mandated JSON Schema for Agent Reasoning. The figure shows the output template (left) and an in-context example of handling a pop-up overlay (right).}
\label{fig:output_prompt_pipeline}
\end{figure*}

\mypara{In-Depth Analysis of PEP.}
\label{sec:pep_analysis_main}
We study PEP's cold-start behavior and entry staleness under app evolution. As shown in Figure~\ref{fig:pep_analysis}(a), even with zero entries the system achieves 53.8\% SR (thanks to PSE and PEC); adding 50 entries boosts SR to 60.4\%, and SR saturates $\sim$500 entries, indicating that a compact pool of recurring patterns suffices. For staleness (Figure~\ref{fig:pep_analysis}(b)), we compare three strategies across five simulated app-update rounds: \textit{No Expiry} degrades SR to 49.6\%, \textit{LRU Eviction} retains 61.5\%, while our \textit{Version-Aware} policy with exponential staleness decay best preserves effectiveness at 65.4\% ($-$3.0\% vs.\ the fresh pool).

\subsection{Details of Diffusion-based Future Layout Generation}
\label{sec:diffusion_details}

For completeness, we outline the training setup of the conditional latent diffusion model used in PSE. Each layout is linearized into a sequence of at most $N$ UI elements $e=(\texttt{type},\texttt{bbox})$ sorted in reading order. Element types are embedded with a learned lookup table, while bounding boxes $(x_{\min},y_{\min},x_{\max},y_{\max})$ are normalized to $[0,1]$ and projected by a linear layer. A Transformer encoder $E_\phi$ maps the ground-truth next layout $L_{t+1}$ to a clean latent $\mathbf{z}_0\in\mathbb{R}^d$, and a Transformer decoder $D_\psi$ reconstructs the ordered element sequence. The encoder--decoder is trained with element-type cross-entropy and bounding-box regression losses. Following latent diffusion~\cite{Rombach_2022_CVPR}, the denoiser $\epsilon_\theta$ is conditioned on $\mathbf{C}=f_\omega(L_t,c_{t,j})$ and optimized as
\begin{equation}
\begin{aligned}
\mathbf{z}_0 &= E_\phi(L_{t+1}), \\
\mathbf{z}_i &= \sqrt{\bar{\alpha}_i}\,\mathbf{z}_0
+ \sqrt{1-\bar{\alpha}_i}\,\boldsymbol{\epsilon}, \\
\mathcal{L}_{\text{diff}}(\theta)
&= \mathbb{E}_{\substack{
(L_t,c_{t,j},L_{t+1})\sim\mathcal{D},\\
i\sim\mathcal{U}\{1,\ldots,N_{\mathrm{train}}\},\,
\boldsymbol{\epsilon}\sim\mathcal{N}(\mathbf{0},\mathbf{I})}}
\left[
\left\|
\boldsymbol{\epsilon}
-\epsilon_\theta(\mathbf{z}_i,i,\mathbf{C})
\right\|_2^2
\right].
\end{aligned}
\end{equation}
Here, $\mathcal{D}$ denotes the trajectory training distribution, and the conditioning vector is obtained by concatenating pooled embeddings of the current layout and concrete executable command, followed by a linear projection. The reconstruction and diffusion objectives are optimized jointly, with the reconstruction terms supervising element types and bounding boxes and $\mathcal{L}_{\text{diff}}$ supervising the conditional latent transition.

We train the model on the training splits of InterfereBench, AndroidControl~\cite{NEURIPS2024_androidcontrol}, and GUI-Odyssey~\cite{2024_gui_odyssey}, using ground-truth next layouts as supervision; held-out evaluation trajectories are excluded as described in Sec.~\ref{sec:impl}. We use $N_{\mathrm{train}}=100$ diffusion timesteps with a cosine noise schedule and optimize with AdamW (learning rate $1\times10^{-4}$, weight decay $0.01$, batch size $128$) and gradient clipping at $1.0$. At inference, we use a deterministic reverse sampler over $N_{\mathrm{sample}}=50$ selected timesteps. The initial latent is sampled from $\mathbf{z}_{N_{\mathrm{train}}}\sim\mathcal{N}(\mathbf{0},\mathbf{I})$ using a fixed per-instance seed; conditional on this initial latent, the reverse trajectory introduces no additional stochastic noise. The resulting latent is decoded by $D_\psi$ into element-type logits and bounding-box offsets, yielding the symbolic layout $\hat{L}_{t+1}$ consumed by the anomaly and reliability estimators in Sec.~\ref{sec:PSE}.

\begin{figure*}[!ht]
\centering
\includegraphics[width=0.9\linewidth]{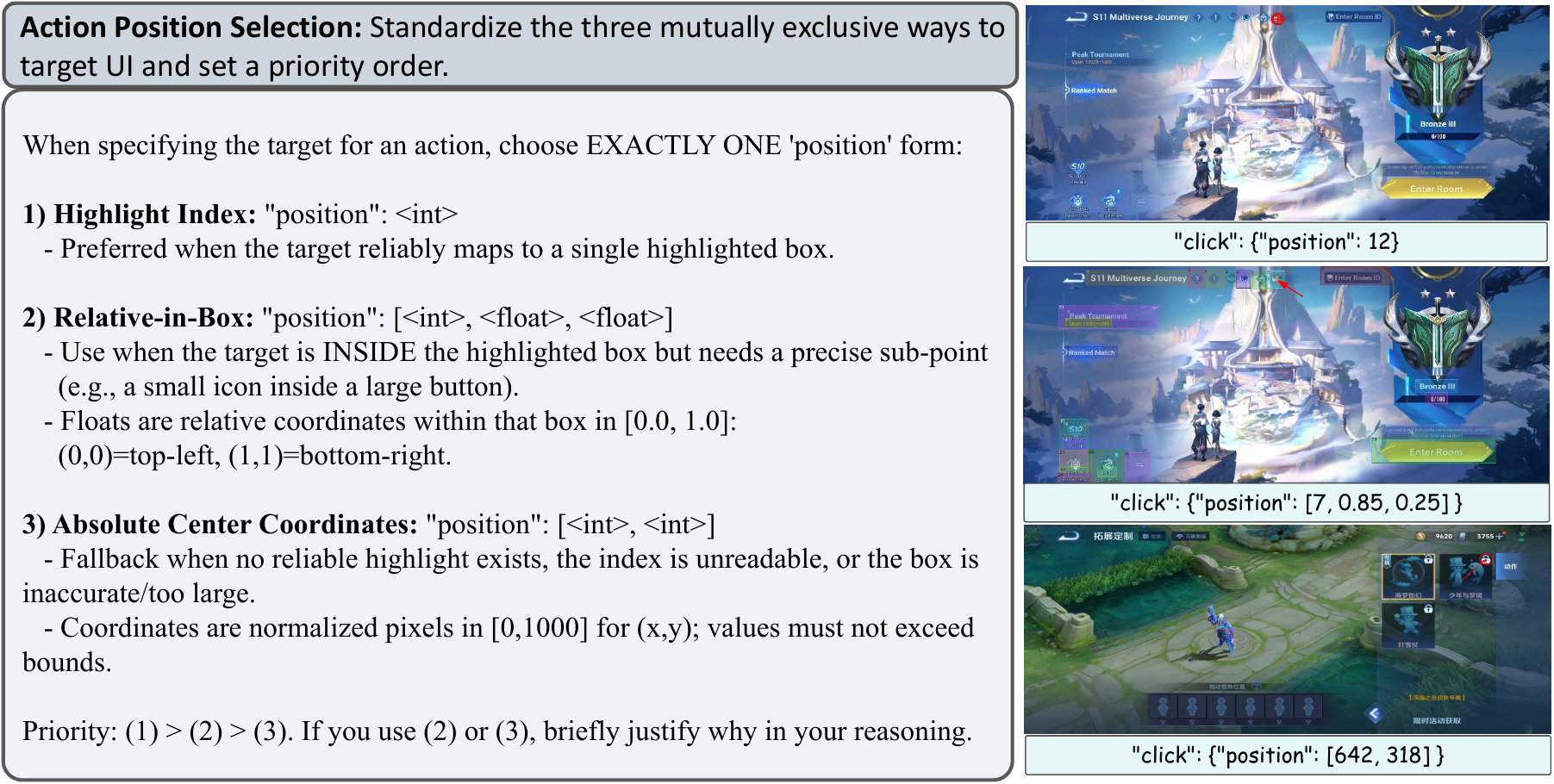} 
\caption{Action commands use three coordinate formats with a default preference of Index $\rightarrow$ Relative-in-Box $\rightarrow$ Absolute. PSE may override this preference when another action-format pair has higher predicted reliability and is not flagged as high-risk.}
\label{fig5:action_template}
\end{figure*}

\begin{figure*}[!t]
\centering
\includegraphics[width=0.9\linewidth]{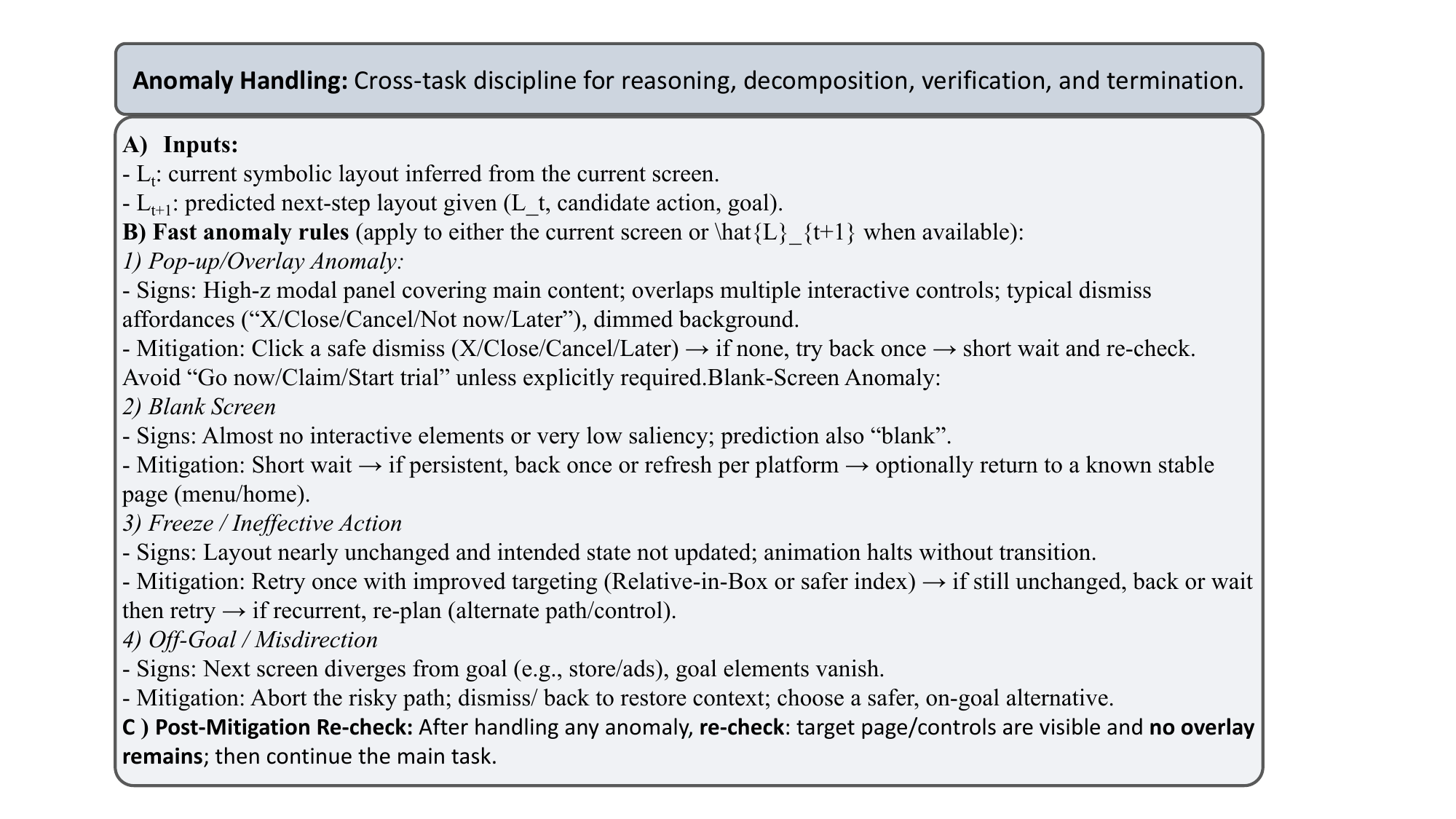} 
\caption{Anomaly Handling Template. The figure illustrates the structured anomaly-handling workflow, including detection rules for pop-ups, blank screens, and freezes (left), and an in-context example of dismissing a pop-up overlay (right).}
\label{fig:Anomaly_handling}
\end{figure*}

\subsection{Hyperparameter Analysis of the Exploration Value}
\label{sec:exploration_value}
We study the single coefficient $\alpha$ that balances novel-control discovery against rare-state probing during exploration. As illustrated in Figure~\ref{fig:alpha_exploration_50steps}, on a 50-step horizon, larger settings (\eg, $\alpha\!\ge\!0.5$) consistently deliver higher cumulative coverage and higher moving-average novelty, indicating faster expansion of the actionable UI space. Very small $\alpha$ emphasizes repeatedly visiting under-explored screens; while this can stabilize early behavior, it sacrifices coverage and slows progress. We observe no significant increase in redundancy within 50 steps, suggesting that short-horizon exploration benefits most from prioritizing discovery. In practice, $\alpha\!\in\![0.5,0.75]$ is a strong operating region that front-loads novel controls without noticeable revisit overhead. For longer horizons or highly volatile apps, an adaptive schedule is preferable: start near $\alpha\!\approx\!0.5$ to stabilize initial navigation, then increase toward $0.75$--$1.0$ as the uncovered-control ratio declines. Overall, $\alpha$ provides an interpretable knob for exploration granularity; tuning (or scheduling) it materially impacts coverage speed and downstream success rates.

\begin{figure*}[!ht]
\centering
\includegraphics[width=0.9\linewidth]{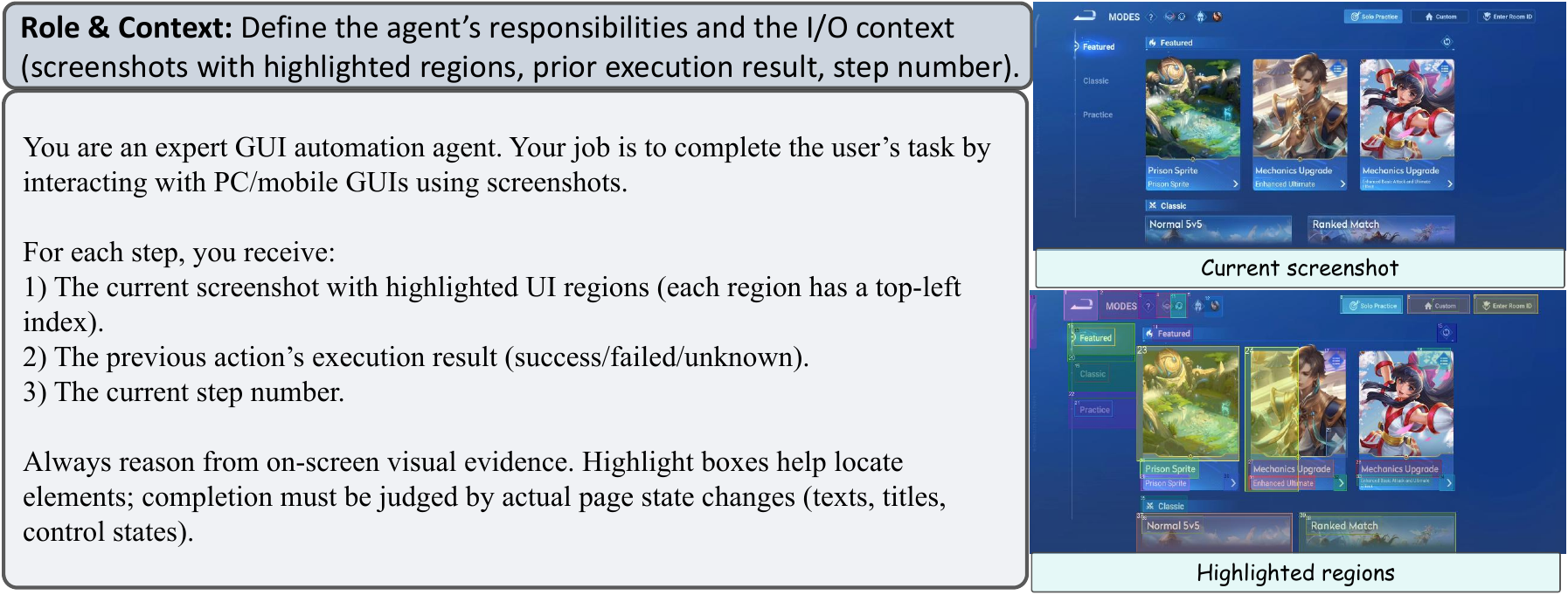} 
\caption{Role and context template. Specifies agent responsibilities and I/O context with indexed screenshots, prior execution results, and step numbers to guide evidence-based task completion.}
\label{fig:role_context}
\end{figure*}

\begin{figure*}[!ht]
\centering
\includegraphics[width=0.94\linewidth]{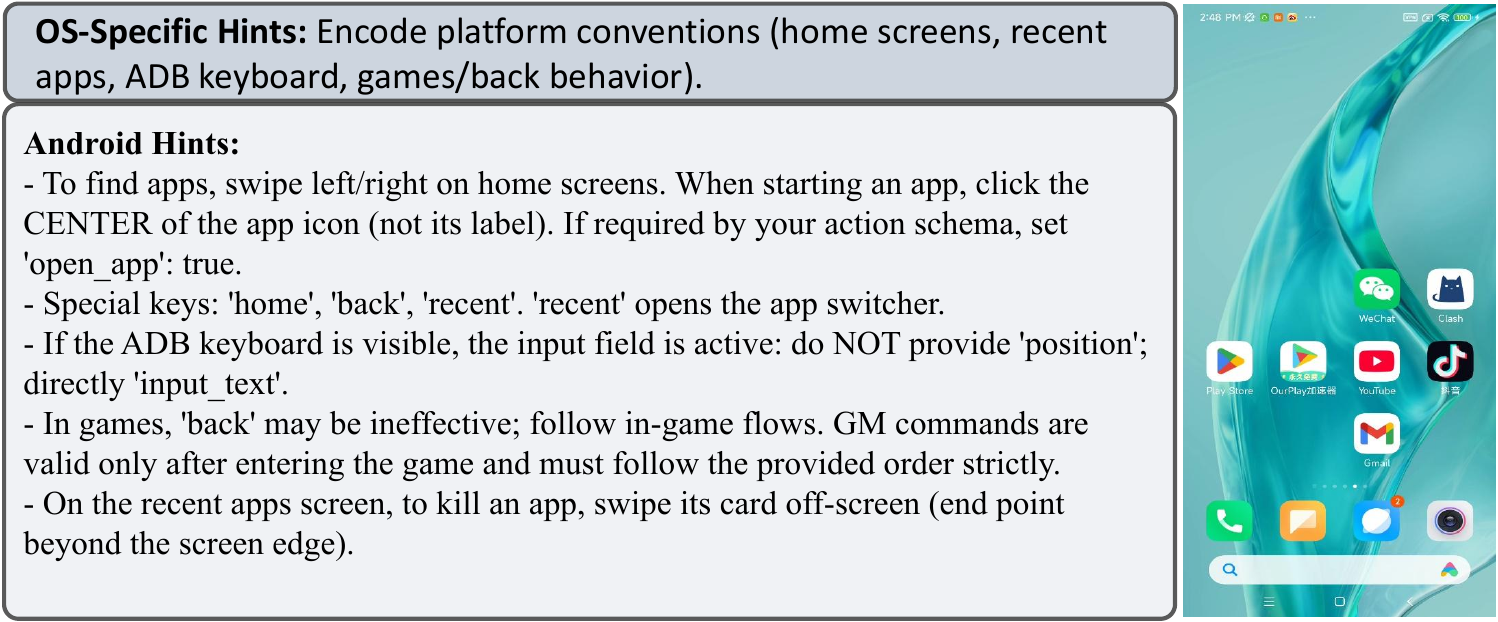} 
\caption{OS-specific action hints. Encodes Android conventions for app access, navigation keys, keyboard input, in-game flows, and app termination to ensure robust, context-aware execution.}
\label{fig:os_hints}
\end{figure*}

\subsection{Parser-Threshold Analysis}
\label{sec:parser_thresholds}
\mypara{Filtering Statistics.}
Stage-1 removes trajectories with excessive length and redundancy. We formalise the self-loop and no-op statistics as:
\begin{equation}
\begin{aligned}
\rho_{\mathrm{loop}} &= \frac{1}{T} \sum_{t=1}^{T}
    \mathbf{1}\!\left[a_t \in \mathcal{A}_{\mathrm{l}}\right]\,
    \frac{\max\!\bigl(0,\,\tau_{\text{c}} - \mathcal{D}_{\text{layout}}(L_t, L_{t+1})\bigr)}{\tau_{\text{c}}}, \\
\rho_{\mathrm{noop}} &= \frac{1}{T} \sum_{t=1}^{T}
    \frac{\max\!\bigl(0,\,\tau_{\text{c}} - \mathcal{D}_{\text{layout}}(L_t, L_{t+1})\bigr)}{\tau_{\text{c}}},
\end{aligned}
\end{equation}
where $T$ is the number of steps, $a_t$ is the action at step $t$, and $L_t$ is the UI layout at step $t$. Here, $\mathbf{1}[\cdot]$ denotes the indicator function, $\mathcal{D}_{\text{layout}}$ is the layout-difference measure derived from the Dice-style similarity in Eq.~\eqref{eq:layout-sim}, $\tau_{\text{c}}$ is the no-op threshold, and $\mathcal{A}_{\mathrm{l}}$ denotes the set of actions annotated as layout-preserving self-loops. A trajectory is pruned as structurally low-quality if either statistic exceeds its preset threshold.

We examine how the Stage-1 pruning thresholds (self-loop ratio and no-op ratio) interact with the no-op cutoff and impact downstream quality, as shown in Fig.~\ref{fig:parser_thresholds_tradeoff}. As the cutoff increases, the acceptance rate drops monotonically across all regimes (\eg, from $\sim$0.60--0.65 at a low cutoff of 0.05 to $\sim$0.20 at 0.45), indicating that more micro-changes are filtered as no-ops.
\textit{Strict} pruning rapidly depresses acceptance (often $<\!0.25$ once the cutoff exceeds $\approx$0.20), and downstream quality declines as data volume becomes the bottleneck.
\textit{Lenient} pruning maintains high acceptance ($>\!0.55$ across most cutoffs) but retains many low-signal segments; the success proxy plateaus or degrades when the cutoff is high (\eg, normalized success $\lesssim$0.55 once the cutoff $\geq$0.35).
By contrast, the \textit{Moderate} regime achieves the best balance in a mid-range cutoff of \textbf{0.25--0.35}: acceptance stays around \textbf{0.35--0.50} while the normalized success proxy peaks around \textbf{0.75--0.85}, yielding the highest harmonic mean of acceptance and success.

\subsection{PSE Prediction Quality Evaluation}
\label{sec:pse_quality}

We evaluate PSE's predicted future layouts on held-out trajectory splits from all three training corpora. Each predicted layout $\hat{L}_{t+1}$ is compared with the ground-truth layout $L_{t+1}$ using one-to-one greedy IoU matching at a threshold of 0.5.

Table~\ref{tab:pse_quality} reports three metrics: (i) \textit{Type Accuracy}---the fraction of matched element pairs whose predicted type label is correct; (ii) \textit{Bbox IoU}---the mean IoU of matched bounding boxes; and (iii) \textit{Element-level F1}---the harmonic mean of element-level precision and recall under IoU$\geq0.5$ matching. The Overall row is computed by pooling all predictions and references from the three test splits before calculating each metric; it is therefore not the arithmetic mean of the three dataset-level percentages. Under this pooled evaluation, PSE obtains 89.4\% type accuracy, 80.2\% mean bbox IoU, and 83.2\% element-level F1. Results on AndroidControl and GUI-Odyssey further show that the predictor transfers across held-out trajectories from multiple GUI corpora; we do not interpret these trajectory-level splits as evidence of app-disjoint generalization.

\begin{table}[!ht]
\centering
\caption{PSE layout prediction quality on held-out trajectory splits. Overall metrics are computed after pooling predictions and references across the three splits.}
\label{tab:pse_quality}
\scriptsize
\setlength{\tabcolsep}{2.5pt}
\begin{tabular}{lccc}
\toprule
\textbf{Dataset} & \textbf{Type Acc. (\%)} & \textbf{Bbox IoU (\%)} & \textbf{Element F1 (\%)} \\
\midrule
InterfereBench  & 91.3 & 82.6 & 85.4 \\
AndroidControl  & 88.7 & 79.1 & 82.3 \\
GUI-Odyssey     & 87.2 & 77.8 & 80.9 \\
\midrule
\textbf{Overall} & \textbf{89.4} & \textbf{80.2} & \textbf{83.2} \\
\bottomrule
\end{tabular}
\end{table}

\mypara{Boundary Case Analysis of Eq.~\eqref{eq:score}.}
The dissimilarity heuristic in Eq.~\eqref{eq:score} assumes that larger layout changes correlate with successful actions. Two edge-case categories violate this assumption: (A) \textit{Success with low $\Delta$layout}---actions like confirming a dialog or dismissing a toast produce correct outcomes yet minimal layout change (8.6\% of test transitions); (B) \textit{Failure with high $\Delta$layout}---accidental taps that trigger page jumps induce large layout shifts despite being incorrect (5.2\%). When relying solely on Eq.~\eqref{eq:score}, the misjudge rates for these two categories are 72.4\% and 68.1\%, respectively. The anomaly-aware weight first filters predictions flagged as severe; PEC's post-execution semantic verification then catches residual errors. With both safeguards, the residual system-level decision errors drop to 18.3\% and 12.7\%. Because the latter verification occurs after execution, these final rates characterize the complete controller rather than PSE prediction alone.

\mypara{Desktop/Web Benchmark: OSWorld.}
OSWorld~\cite{NEURIPS2024_5d413e48} contains 369 tasks involving real web and desktop applications, OS file I/O, and cross-application workflows. We follow the screenshot-only protocol and the 15-/50-step action budgets used by UI-TARS~\cite{qin2025_uitars}. Baseline scores in Table~\ref{tab:osworld} are reported under this protocol; PrecogUI reaches 20.7\% and 27.5\% SR under the 15- and 50-step budgets, respectively. It exceeds UI-TARS-7B-DPO by 2.0 points at 15 steps and UI-TARS-72B-DPO by 2.9 points at 50 steps, while UI-TARS-72B-DPO remains 2.0 points higher under the 15-step budget.

\begin{table}[!ht]
\centering
\caption{Success rate (\%) on OSWorld under the UI-TARS screenshot-only protocol.}
\label{tab:osworld}
\footnotesize
\setlength{\tabcolsep}{5pt}
\begin{tabular}{lcc}
\toprule
\textbf{Method} & \textbf{15-step} & \textbf{50-step} \\
\midrule
GPT-4o~\cite{openai2024gpt4ocard}                & 5.0  & -- \\
CogAgent-9B~\cite{Hong_2024_CVPR}                & 8.1  & -- \\
OS-Atlas-7B~\cite{DOS-ATLAS}                     & 14.6 & -- \\
Aguvis-72B~\cite{xu2025aguvis}                   & 17.0 & -- \\
UI-TARS-7B-SFT~\cite{qin2025_uitars}             & 17.7 & -- \\
UI-TARS-7B-DPO~\cite{qin2025_uitars}             & 18.7 & -- \\
UI-TARS-72B-SFT~\cite{qin2025_uitars}            & 18.8 & -- \\
UI-TARS-72B-DPO~\cite{qin2025_uitars}            & \textbf{22.7} & 24.6 \\
\midrule
\textbf{PrecogUI}                                & 20.7 & \textbf{27.5} \\
\bottomrule
\end{tabular}
\end{table}

\begin{figure*}[!ht]
\centering
\includegraphics[width=0.99\linewidth]{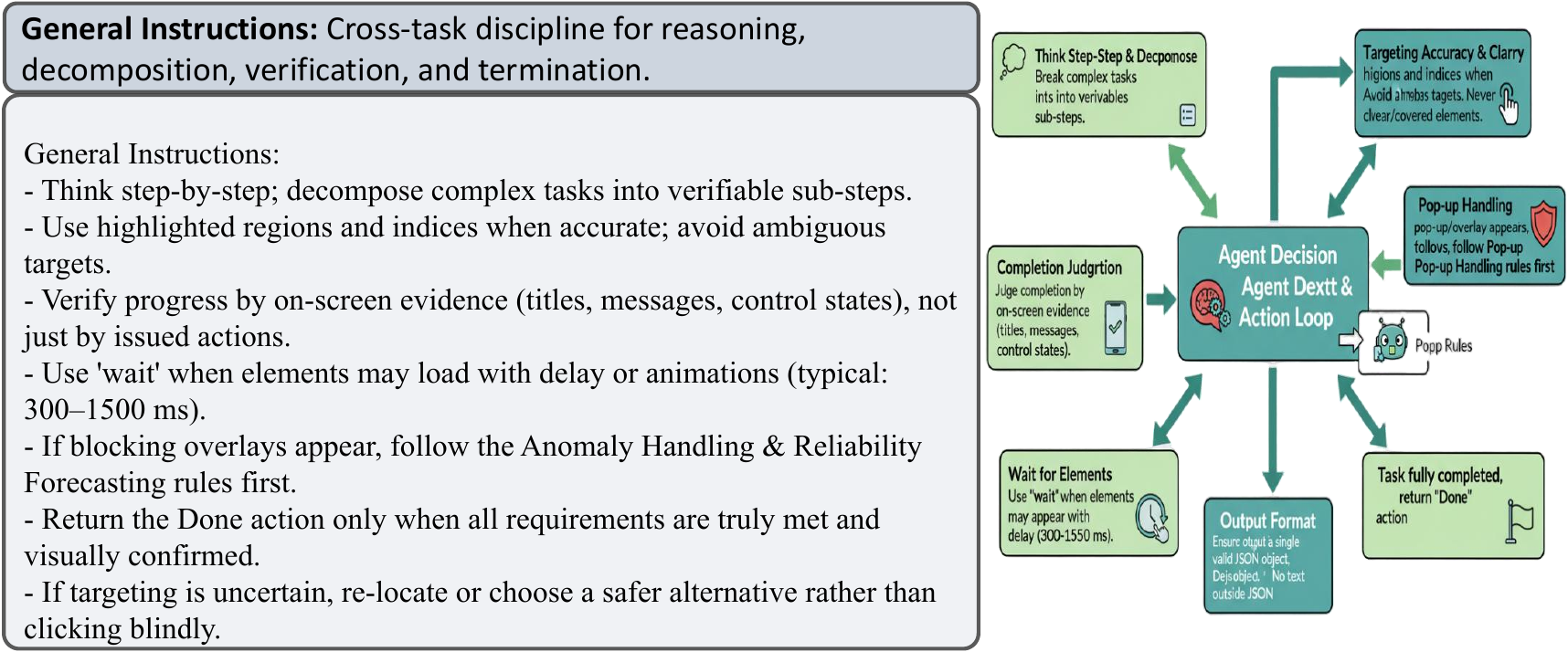} 
\caption{General instruction template. Defines structured reasoning, precise targeting, verification, controlled waiting, and disciplined termination to ensure robust, evidence-driven task execution.}
\label{fig:general_instructions}
\end{figure*}

\begin{figure*}[!ht]
\centering
\includegraphics[width=0.99\linewidth]{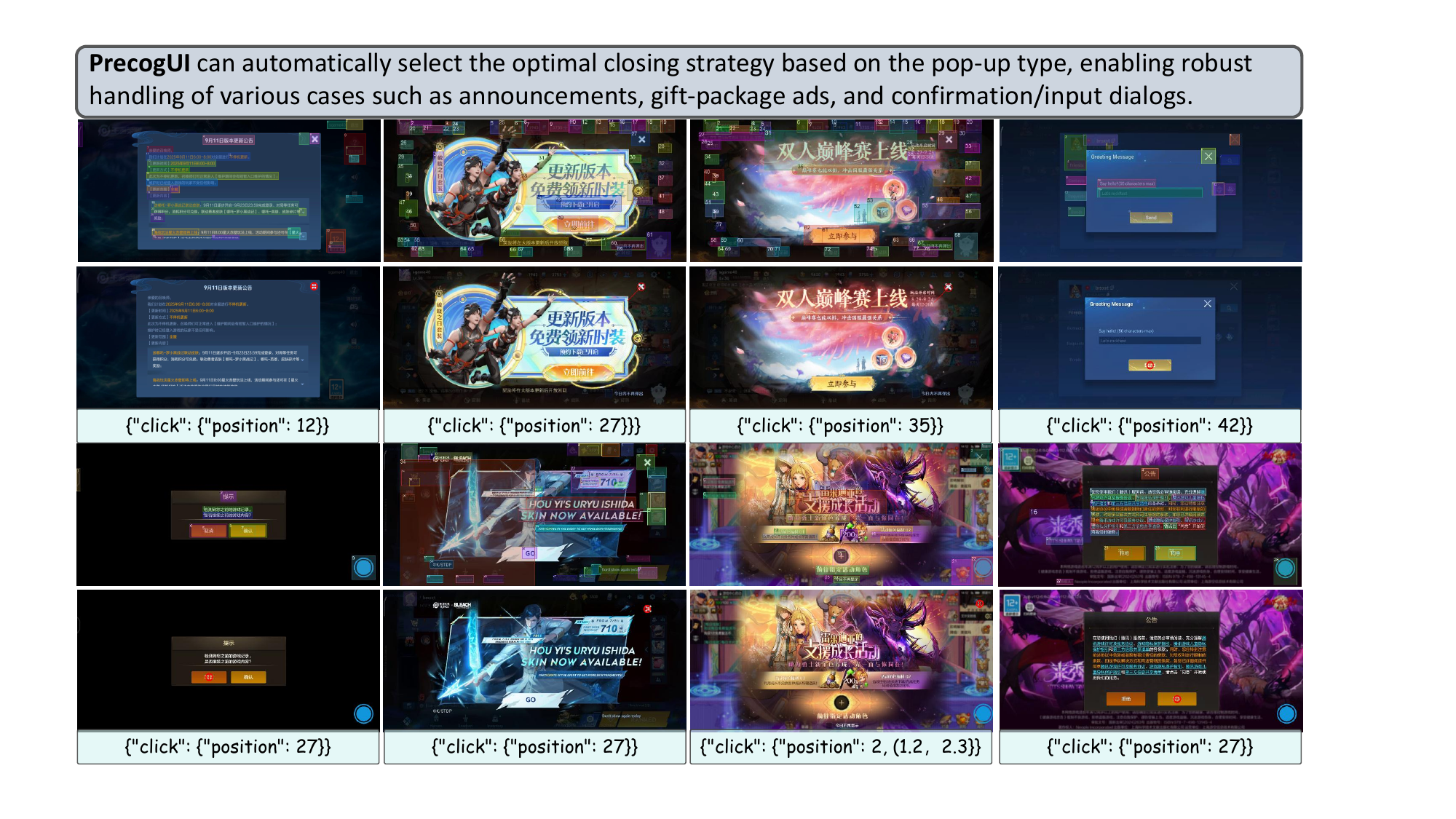} 
\caption{Apps pop-up handling. A type-aware policy combined with hierarchical position selection (Index $\rightarrow$ Relative-in-Box $\rightarrow$ Absolute); the figure presents concrete dismissal commands for diverse pop-up cases.}
\label{fig:inapp_popups}
\end{figure*}

\section{Prompts in Automated Pipeline}
\label{sec:prompt}

\subsection{Output Format Structure Template}
\label{sec:output_Format}
As illustrated in Figure~\ref{fig:output_prompt_pipeline}, our \texttt{Deep Think \& Decision} mechanism is governed by a mandated JSON schema that structures the agent's output. This schema enforces a rigorous, multi-stage reasoning process through several key fields: \texttt{Historical\_status} for visual verification of the previous action's outcome, severing reliance on potentially noisy execution logs; \texttt{import\_contents} for grounding the agent's awareness in the current UI context; \texttt{think} for articulating a step-by-step causal rationale; \texttt{progress} and \texttt{next\_goal} for explicit task decomposition and forward planning; and finally \texttt{action}, which specifies the precise, parameterized command for environmental actuation (\eg, via index-based coordinates). Crucially, the schema's emphasis on populating fields like \texttt{Historical\_status} based \textit{solely} on visual evidence establishes a tight closed-loop verification system. This structured output thereby functions as a transparent and auditable interface between the agent's cognitive deliberation and its concrete actions within the GUI environment.

\begin{figure*}[!ht]
\centering
\includegraphics[width=0.98\linewidth]{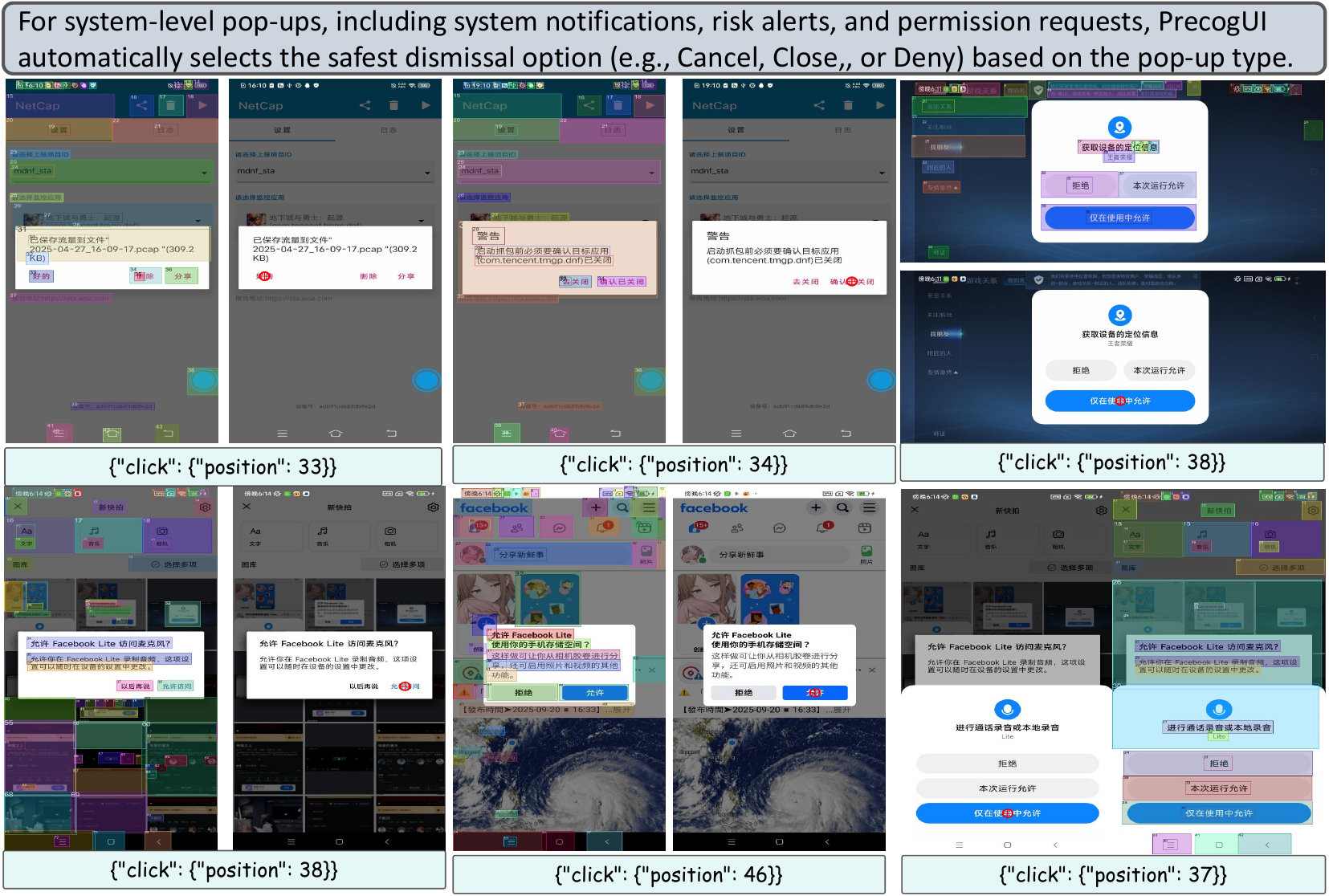} 
\caption{System pop-up handling. A type-aware policy selects safe dismissal actions and executes them with index-prioritized targeting; the figure shows concrete commands for notifications, risk alerts, and permission requests.}
\label{fig:system_pop_up}
\end{figure*}

\subsection{Action Selection Template}
\label{sec:Action_Selection_Template}
To ensure robust action grounding, we define three coordinate formats. \textbf{(1) Highlight Index} targets an element through a semantic identifier and is used as the default when the index is stable. \textbf{(2) Relative-in-Box} specifies a sub-point within an indexed element and is useful when the desired target is only part of a larger control. \textbf{(3) Absolute Coordinates} provide a normalized fallback when semantic indexing is unavailable. The order Index $\rightarrow$ Relative-in-Box $\rightarrow$ Absolute is a default preference, not a hard constraint. At each step, PEC evaluates the available action-format pairs with PSE, removes candidates whose predicted risk exceeds the safety threshold, and executes the remaining pair with the highest relative reliability score. Consequently, Relative-in-Box or Absolute may be selected ahead of Index when the current layout makes the default format unreliable.

\subsection{Anomaly Handling Template}
\label{sec:Workflow_Exception_Handling}
As shown in Figure~\ref{fig:Anomaly_handling}, we frame anomaly handling as a concise, cross-task routine over prediction and verification. Given the current layout $L_t$ and forecast $\hat{L}_{t+1}$, fast rules classify the risk, after which PEC selects a mitigation: (i) Pop-up/Overlay---dismiss via safe affordances (\texttt{X/Close/Cancel/Later}); (ii) Blank Screen---wait briefly, then \texttt{Back}/refresh or return to a stable hub; (iii) Freeze/Ineffective Action---retry once with safer targeting, else Back/re-plan; (iv) Off-Goal/Misdirection---abort the path and restore on-goal context. A compulsory re-check gates progress after mitigation.

\begin{figure*}[!ht]
\centering
\includegraphics[width=0.98\linewidth]{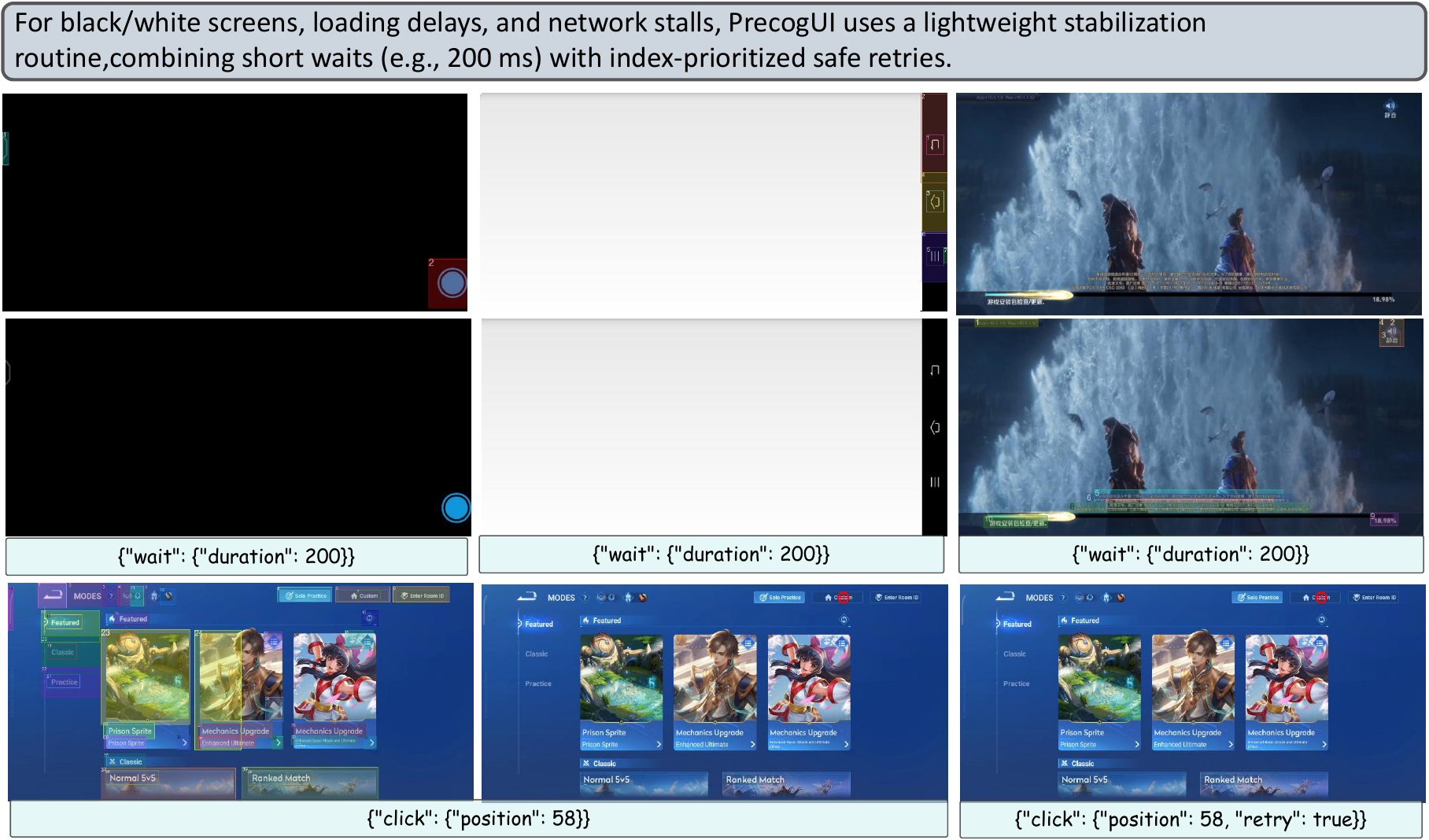} 
\caption{Environment disturbances. A lightweight routine---short waits plus index prioritized safe retries stabilizes black/white screens, delayed loads, and network stalls; the figure shows concrete \texttt{wait} and \texttt{retry} commands for representative cases.}
\label{fig:Environment_Perturbation}
\end{figure*}

\subsection{Role and Context Template}
\label{sec:role_context}
To structure the agent's operational context, we define a clear set of responsibilities and a standardized input format for each reasoning step. As illustrated in Figure~\ref{fig:role_context}, the agent is prompted with persona as an expert GUI automation agent. For each step, it receives a tripartite input: (1)~the current screenshot augmented with indexed, highlighted bounding boxes over interactable elements; (2)~feedback on the execution status (\eg, success or failure) of the prior action; and (3)~the current temporal step index. Crucially, the agent is explicitly instructed to ground its reasoning \textit{solely on visual evidence}, judging task progression based on observable changes in the UI state rather than uncritically accepting the programmatic execution status. This mandate establishes a tight, closed-loop visual verification process for all decision-making.

\subsection{OS-Specific Hints}
\label{sec:os_hint}
As shown in Figure~\ref{fig:os_hints}, we encode platform conventions into structured hints that guide robust action execution on Android. These rules address common UI operations and context-sensitive behaviors: (i) app launching via centered icon clicks with optional \texttt{open\_app} flag; (ii) special system keys such as \texttt{home}, \texttt{back}, and \texttt{recent} for navigation control; (iii) text input handling by directly invoking \texttt{input\_text} when the ADB keyboard is active, avoiding redundant position specifications; (iv) game-specific flows where the \texttt{back} key may be ineffective, requiring strict adherence to in-game command order; and (v) app termination through swipe-off gestures in the recent-apps screen. Collectively, these hints ground agent actions in OS-level semantics, reducing execution ambiguity and improving cross-context stability.

\begin{figure*}[!ht]
\centering
\includegraphics[width=0.98\linewidth]{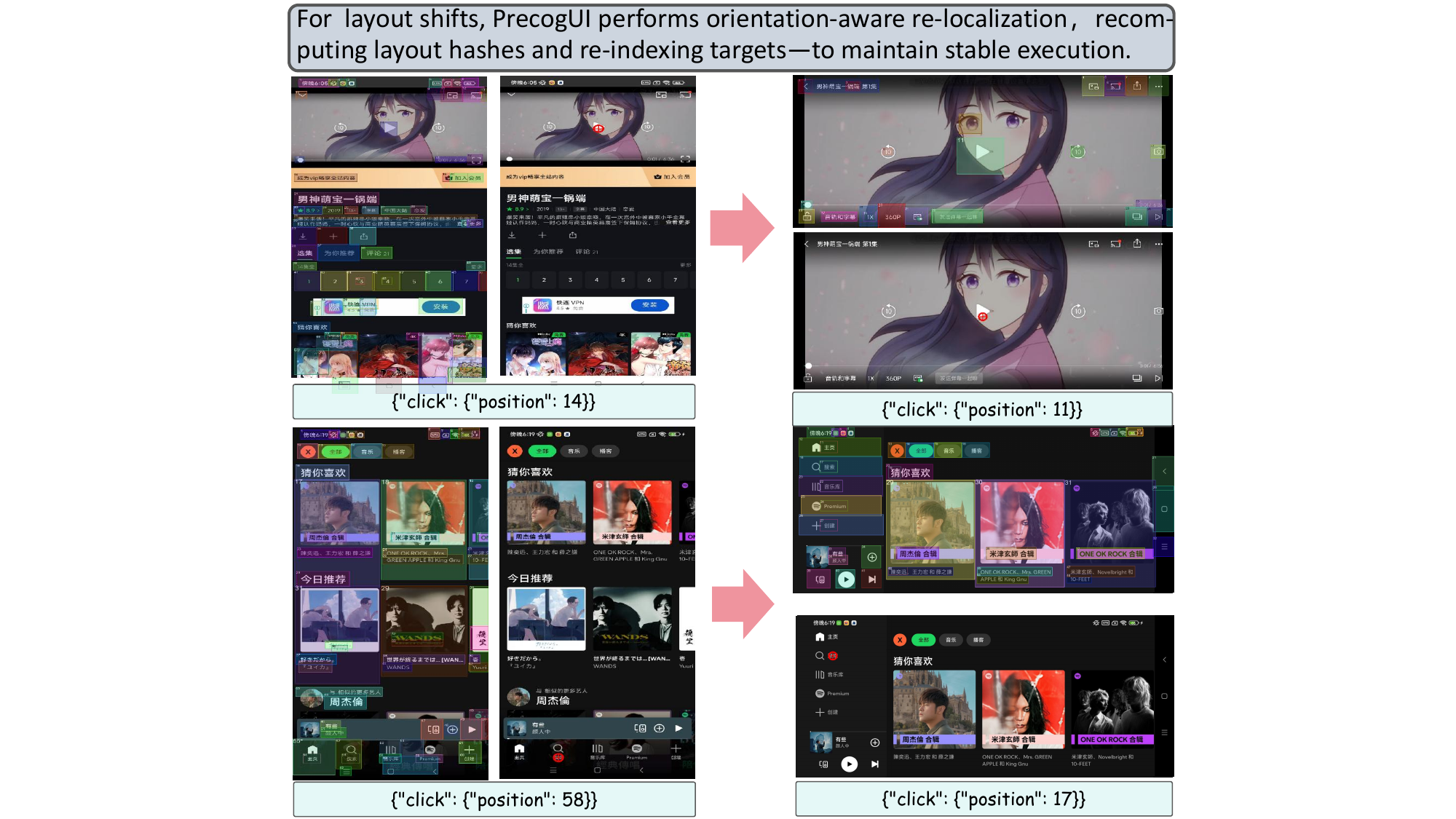} 
\caption{Layout-shift handling. PrecogUI rebuilds layout hashes and re-indexes targets under portrait/landscape transitions, executing with index-first targeting; the figure shows before/after screens with preserved action intent.}
\label{fig:layout_shift}
\end{figure*}

\subsection{General Instructions}
\label{sec:general_instructions}
As shown in Figure~\ref{fig:general_instructions}, this template encodes cross-task discipline for structured reasoning and verifiable execution. It emphasizes (i) step-by-step task decomposition into checkable sub-steps; (ii) precise targeting using highlighted regions or indices while avoiding ambiguous clicks; (iii) progress verification strictly by on-screen evidence such as titles, messages, or control states; (iv) controlled waiting to accommodate delays or animations; (v) fallback to anomaly-handling rules when overlays appear; and (vi) termination only after explicit visual confirmation of success. When targeting remains uncertain, the agent is required to re-locate or choose safer alternatives, ensuring robustness against cascading errors. Collectively, these rules establish a disciplined action loop where correctness validation precedes task advancement.

\section{Qualitative Analysis}
\label{sec:Qualitative}

\subsection{Apps Pop-up Handling}
\label{sec:Apps_Pop-up}
As shown in Figure~\ref{fig:inapp_popups}, we deploy a type-aware policy that closes in-app pop-ups while preserving task context. The controller first classifies the pop-up---(i) announcement/notice panels, (ii) gift-package ads, (iii) event promotions, or (iv) confirmation/input dialogs---and selects the safest dismiss affordance. Execution follows our hierarchical position schema: prioritize element \emph{indices} for \texttt{X/Close/Cancel/Later}; degrade to \emph{Relative-in-Box} when the target is a sub-control; and use \emph{normalized absolute} coordinates only when indexing is unreliable. Each thumbnail shows the predicted command (index or relative point) rendered beneath the image; progress continues only after the overlay is visually cleared.

\subsection{System-Level Pop-up Handling}
\label{sec:System-Level Pop-up}
As shown in Figure~\ref{fig:system_pop_up}, we handle OS-mediated interruptions---system notifications, risk alerts, and permission requests---via a type-conditioned, safety-first policy. The controller classifies the pop-up and selects the safest affordance (\eg, \texttt{Cancel/Close}, \texttt{Allow only while in use}, \texttt{Deny}). Execution uses our hierarchical position scheme, prioritizing element \emph{indices} and backing off to \emph{Relative-in-Box} or normalized \emph{Absolute} coordinates only when indexing is unreliable. Each panel displays the issued command (primarily index clicks), and progress resumes only after the overlay is visually cleared to preserve task context.

\subsection{Environment Perturbation Handling}
As shown in Figure~\ref{fig:Environment_Perturbation}, we address environment-level disturbances (black/white screens, loading delays, and network stalls) with a lightweight stabilization routine. Detection relies on low-saliency/blank frames, near-identical consecutive layouts, or stalled progress indicators. Mitigation is minimal yet effective: inject a short wait (\eg, 200\,ms) to absorb transient transitions, then issue a single index-prioritized safe retry of the previous action; progress resumes only after visual evidence of recovery, otherwise control is escalated to the general anomaly rules.

\mypara{Layout-Shift Perturbations.}
As shown in Figure~\ref{fig:layout_shift}, we address orientation/gravity--induced reflows (portrait $\leftrightarrow$ landscape) with an orientation-aware re-localization routine. Upon detecting a layout shift (aspect-ratio change and index invalidation), the agent reconstructs the symbolic layout hash, re-indexes targets, and remaps the current goal to the new arrangement by type/text cues. Execution then follows the hierarchical position policy (Index $\rightarrow$ Relative-in-Box $\rightarrow$ Absolute), and progress is gated by visual re-check to ensure the intended control is active after rotation.

\section{PEC Algorithm Pseudocode}
\label{sec:pec_algorithm}

Algorithm~\ref{alg:pec} summarizes the inference-time control flow of PEC and complements the module description in Sec.~\ref{sec:PEC}. PEC first handles anomalies already visible in the current state by retrieving a remedy from PEP or invoking the structured anomaly-diagnosis policy. Otherwise, it generates candidate action-format pairs, filters those predicted to be high-risk by PSE, and executes the highest-scoring safe pair. Each outcome is then verified using visual and semantic feedback; failed pairs are temporarily blocked, while unexpected transitions trigger recovery or episode restart before candidate generation resumes.

\begin{algorithm}[!ht]
\caption{Pre-cognitive Execution Controller}
\label{alg:pec}
\begin{algorithmic}[1]
\Require Goal $g$, current state $s_t$, anomaly memory $M_a$, anomaly rules $\mathcal{R}_{\text{anom}}$
\Ensure A verified action-format pair $(a,r)$ or an anomaly-handling action

\Statex \textit{// -- Stage 1: Pre-cognitive Anomaly Checks --}
\State $\ell_t \gets \text{Layout}(s_t)$
\If{$M_a.\text{QueryByLayout}(\ell_t)$ returns a remedy $a_{\text{handle}}$}
    \State \Return $a_{\text{handle}}$
\EndIf
\If{$\text{CurrentStateAnomaly}(s_t;\mathcal{R}_{\text{anom}})$} \Comment{Current-state check, not action forecasting}
    \State \Return $\pi_{\text{LLM}}(s_t, \mathcal{T}_{\text{anomaly}})$ \Comment{Autonomous diagnosis for novel anomaly}
\EndIf

\Statex \textit{// -- Stage 2: Iterative Execution and Recovery Loop --}
\State $\mathcal{C}_t \gets \text{MLLM.GenerateCandidatePairs}(s_t, g)$
\State $\mathcal{F}_t \gets \emptyset$ \Comment{Initialize temporary taboo list}
\While{$\mathcal{C}_t \setminus \mathcal{F}_t$ is not empty}
    \State $\mathcal{C}_t^{\text{safe}} \gets
    \{(a,r)\in\mathcal{C}_t\setminus\mathcal{F}_t:
    w(\text{PSE.ForecastLayout}(s_t,a,r))>0\}$
    \If{$\mathcal{C}_t^{\text{safe}}$ is empty}
        \State \Return $\pi_{\text{LLM}}(s_t, \mathcal{T}_{\text{anomaly}})$
    \EndIf
    \State $(a^*,r^*) \gets \arg\max_{(a,r) \in \mathcal{C}_t^{\text{safe}}} s(a,r)$
    \State \textbf{execute} $(a^*,r^*)$; \textbf{observe} new state $s_{t+1}$

    \If{$\text{VerifySuccess}_{\text{layout+semantic}}(s_t, (a^*,r^*), s_{t+1})$}
        \State \Return $(a^*,r^*)$ \Comment{\textbf{Success}: terminate step}
    \EndIf

    \State $\mathcal{F}_t \gets \mathcal{F}_t \cup \{(a^*,r^*)\}$
    \If{$\mathcal{D}_{\text{layout}}(\text{Layout}(s_t), \text{Layout}(s_{t+1})) < \tau_{\text{c}}$} \Comment{Stagnation}
        \State \textbf{continue}
    \Else \Comment{Unexpected Transition}
        \State $s_t \gets \Call{RecoverOrRestart}{}$
        \State $\mathcal{C}_t \gets \text{MLLM.GenerateCandidatePairs}(s_t, g)$
        \State $\mathcal{F}_t \gets \emptyset$ \Comment{Discard taboos tied to the failed state}
    \EndIf
\EndWhile

\State \Return $\pi_{\text{LLM}}(s_t, \mathcal{T}_{\text{anomaly}})$ \Comment{Final diagnosis if all candidates fail}
\end{algorithmic}
\end{algorithm}

\section{Additional Discussions}
\label{sec:discussion}

Forecasting future layouts is central to PrecogUI: look-ahead turns reactive "observe--act" behavior into risk-aware planning that preempts pop-ups, freezes, and off-goal drifts, improving long-horizon stability. However, timeliness is a key constraint. Pre-execution simulation and verification add latency and compute, which can be costly for real-time use or very long tasks. In addition, experience priors can become stale as apps update; outdated remedies hurt reliability unless memory is refreshed. Future work should adopt lightweight, anytime forecasting and drift-aware memory maintenance to preserve the gains of look-ahead without sacrificing responsiveness.

\subsection{Cross-Device Grounding and Mobile Navigation Analysis}
\label{sec:cross_device}

PrecogUI represents UI elements with normalized coordinates and structured \texttt{(type,bbox)} layouts, reducing its dependence on any single screen resolution. The Desktop and Web subsets of ScreenSpot provide direct evidence for cross-form-factor grounding: as reported in Table~\ref{tab:ground_performance}, PrecogUI obtains 97.5\% text and 82.2\% icon accuracy on Desktop, and 94.6\% text and 91.7\% icon accuracy on Web. These results support transfer of the grounding component across mobile, desktop, and web screenshots, but they do not by themselves establish end-to-end task completion on desktop or web environments.

AndroidControl-High and GUI-Odyssey evaluate end-to-end navigation in mobile application environments. PrecogUI achieves 76.4\% SR on AndroidControl-High and 89.1\% on GUI-Odyssey, as shown in Table~\ref{tab:Navigation_performance}. OSWorld provides complementary end-to-end evidence on desktop and web applications under a screenshot-only protocol, where PrecogUI obtains 20.7\% and 27.5\% SR with 15- and 50-step budgets (Table~\ref{tab:osworld}). We therefore interpret the evidence conservatively: ScreenSpot measures cross-form-factor grounding, AndroidControl and GUI-Odyssey measure mobile navigation, and OSWorld supplies an initial desktop/web end-to-end evaluation. Broader testing with additional interactive desktop and web protocols remains future work.

\subsection{Rollback Scope and Empirical Statistics}
\label{sec:rollback_analysis}

\mypara{Scope and Feasibility.}
The recovery branch in Sec.~\ref{sec:PEC} and Algorithm~\ref{alg:pec} is restricted to reversible evaluation tasks. We distinguish three operations that were previously grouped under the term rollback. First, \textit{navigation recovery} uses reversible UI actions such as Back, Close, or Home and then visually verifies whether the last stable task context has been recovered. Second, \textit{episode restart} reinitializes a benchmark task from its known starting state when navigation recovery fails. Third, \textit{checkpoint restoration} is used only in emulator environments that expose an explicit state-checkpoint API. We do not claim that an arbitrary physical-device application state can be restored after an external side effect.

\mypara{Snapshot Storage and Restoration.}
Before executing a candidate, PEC stores an \textit{observation record} containing the last confirmed screenshot, its parsed layout, the selected action-format pair, and the verification result. This record supports diagnosis and re-planning but is not itself a restorable application state. When an emulator checkpoint is available, the environment state is restored through the benchmark interface. Otherwise, PEC attempts navigation recovery and falls back to an episode restart if the stable context cannot be re-established. After any successful recovery, PEC regenerates candidate action-format pairs from the recovered state rather than reusing predictions made for the failed state.

\mypara{Empirical Usage Statistics.}
To quantify the practical impact of recovery, we report usage statistics across three long-horizon evaluation splits:

\begin{table*}[!ht]
\centering
\caption{Recovery usage statistics across long-horizon evaluation splits. ``Episodes w/ recovery'' and ``Episodes requiring restart'' are percentages of all evaluated episodes; ``Success after recovery'' is computed only within episodes where recovery was triggered.}
\label{tab:rollback_stats}
\footnotesize
\begin{tabular}{lccc}
\toprule
\textbf{Dataset} & \textbf{Episodes w/ recovery} & \textbf{Success after recovery} & \textbf{Episodes requiring restart} \\
\midrule
AndroidControl-High & 12.4\% & 69.7\% & 1.8\% \\
GUI-Odyssey         & 15.3\% & 73.1\% & 2.4\% \\
InterfereBench      & 18.6\% & 71.2\% & 2.7\% \\
\bottomrule
\end{tabular}
\end{table*}

As shown in Table~\ref{tab:rollback_stats}, recovery is triggered in 12.4--18.6\% of all evaluated episodes. Among those episodes, 69.7--73.1\% resume successfully after navigation recovery or checkpoint restoration. Episode restarts account for 1.8--2.7\% of all evaluated episodes. These statistics quantify recovery within our controlled evaluation protocol; they do not imply that irreversible external side effects can be undone.

\mypara{Handling Irreversible Operations.}
When PEC repeatedly fails from the same stable state, it first retries with alternative safe action-format pairs and then invokes navigation recovery. If the task context cannot be recovered, the evaluation episode is restarted from its initial state. The current PSE detects visual and transition anomalies; it is not a semantic authorization mechanism for transactions or other irreversible actions. Accordingly, our autonomous evaluation excludes operations that can create real-world side effects, such as purchases or message transmission. A deployed system should require explicit user confirmation and application-specific safeguards for such actions. End-to-end protection for irreversible operations remains outside the present scope.

\end{document}